\documentclass[aps,prx,reprint,superscriptaddress,longbibliography,nofootinbib,floatfix]{revtex4-2}
\usepackage{amsmath,amssymb,amsthm,mathtools}
\usepackage{graphicx}
\usepackage[usenames,dvipsnames,svgnames]{xcolor}
\usepackage[hidelinks]{hyperref}
\usepackage{bm}

\hypersetup{
  pdftitle={Symmetry in Time: A Representation Theory of Linear Recurrent Memory},
  pdfauthor={Anif N. Shikder, Ramit Dey, J\'an Min\'a\v{c}, Lyle E. Muller},
  pdfkeywords={state space models, state tracking, representation theory, quaternions},
}

\newcommand{\HH}{\mathbb{H}}
\newcommand{\CC}{\mathbb{C}}
\newcommand{\RR}{\mathbb{R}}
\newcommand{\ZZ}{\mathbb{Z}}

\theoremstyle{plain}
\newtheorem{theorem}{Theorem}
\newtheorem{proposition}{Proposition}
\newtheorem{corollary}{Corollary}
\newtheorem{lemma}{Lemma}
\theoremstyle{definition}
\newtheorem{definition}{Definition}

\usepackage{xcolor}

\begin{document}

\title{Symmetry in Time:\\ A Representation Theory of Linear Recurrent Memory}

\author{Anif N. Shikder}
\thanks{ashikder@uwo.ca}
\affiliation{Fields Lab for Network Computation, Fields Institute, Toronto ON, Canada}

\author{Ramit Dey}
\affiliation{Fields Lab for Network Computation, Fields Institute, Toronto ON, Canada}

\author{Ján Mináč}
\affiliation{Department of Mathematics, Western University, London ON, Canada}
\affiliation{Fields Lab for Network Computation, Fields Institute, Toronto ON, Canada}

\author{Lyle E. Muller}
\affiliation{Department of Neuroscience, University of Texas at Dallas, Richardson TX, United States
}
\affiliation{Fields Lab for Network Computation, Fields Institute, Toronto ON, Canada}

\begin{abstract}
State space models (SSMs) are an emerging alternative to Transformers. SSMs offer linear-time inference and parallel training, but most implementations make the parallel scan efficient by making the recurrence diagonal: each channel of the hidden state is multiplied by an input-dependent real or complex number. These numbers commute, so the state records which inputs occurred but not their order. We formalize this with representation theory, which explains and fixes the state-tracking failures previously reported for SSMs. The state update defines a linear representation of the monoid of input sequences, and we show that an SSM tracks a group at every sequence length exactly when that group is a quotient of the group generated by its transition matrices. Two consequences follow. Diagonal real- and complex-valued SSMs can track only commutative groups, while the transition matrices of any SSM that can track a group must split into independent blocks over the real, complex, and quaternion numbers. We then introduce HyperSSM, which adds quaternion channels while keeping linear-time inference and parallel training. Trained only on final answers, HyperSSM solves the group tasks of recent state-tracking benchmarks exactly with at most five state numbers. This includes $S_5$ with 32 generators, where previous SSMs are at chance, and HyperSSM generalizes to sequences 25 times longer than those seen in training. Because the theory says what the transitions must encode, the learned weights of the linear recurrence can be interpreted directly. We demonstrate this by recovering the calibration of a real gyroscope, the rotations of imperfect qubit gates and the geometry of protein backbones.
\end{abstract}

\maketitle

\section{Introduction}

Transformers are a central machine learning architecture for language and
sequence processing tasks~\cite{vaswani}. They compare every token in a sequence with
every other token, so their memory and computation grow quadratically with
the length of the sequence. State space models (SSMs) have emerged as an
alternative that scales linearly~\cite{gu_s4,orvieto_lru,gu_mamba,dao_mamba2,mamba3}.
In an SSM, information about past inputs is contained in the activity state of the recurrent layer, which is updated at each step by a linear recurrence that can be computed in parallel during training and step by step during inference. SSMs now match transformers on
many sequence modelling benchmarks, and they serve as the backbone, alone or
combined with attention, of deployed language
models~\cite{jamba,nemotronh,falconh1}. As these systems are asked to follow
longer interactions, execute programs and act as agents, a basic question
becomes pressing: can their memory keep track of events in the order in
which they occurred?

\begin{figure*}[t]
\centering
\includegraphics[width=\textwidth]{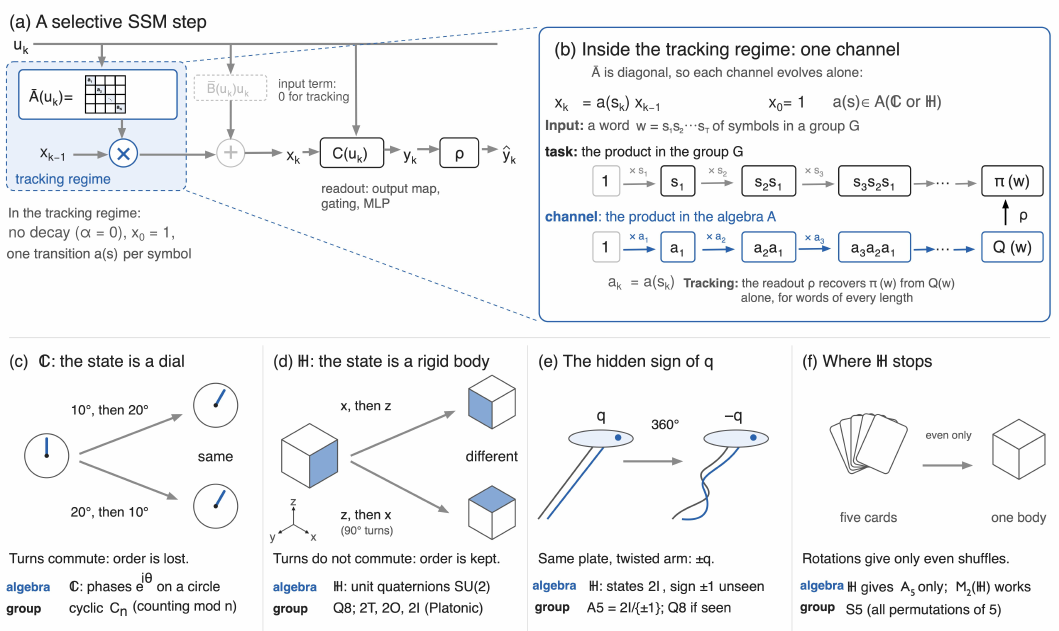}
\caption{\textbf{Order in a linear memory.} (a) One step of a selective
SSM; shaded, the tracking regime (no decay, $x_0=1$, one transition per
symbol). (b) After a word $w$, a channel holds the ordered product $Q(w)$; it
tracks the group if a continuous readout recovers $\pi(w)$ from $Q(w)$ at
every length. (c) A complex channel stores a phase; turns add, and order is
lost. (d) A quaternion channel stores a 3D orientation; $90^\circ$ turns
about $x$ and $z$ in either order give different states. (e) $q$ and $-q$
are the same rotation; a sign-invariant readout tracks $A_5$, a sign-aware
one tracks $Q_8$. (f) $\HH$ tracks $A_5$ but not $S_5$; $M_2(\HH)$ tracks
$S_5$.}
\label{fig:intuition}
\end{figure*}

In early SSMs, the linear recurrence does not depend on the input, so
shifting the input in time shifts the hidden state by the same amount:
before any nonlinear readout, the recurrence acts as a convolution. Just
as the discrete-time Fourier transform diagonalizes any discrete
convolution, such a recurrence is diagonalized by discrete exponentials
$\lambda^k$, where the $\lambda$ are the eigenvalues of its transition
matrix. Because the recurrence multiplies by the same matrix at every
step, and powers of one matrix commute, each of these modes evolves
independently and needs only a single number $\lambda$. In general this
number is complex: a real eigenvalue can only make its mode decay or flip
sign, whereas a complex one also rotates it, storing a phase that lets
the mode respond to periodic structure and count modulo $n$. The hidden
state is therefore a collection of independent real or complex numbers,
each decaying, and possibly rotating, at its own rate. Viewed on a ring of
nodes, these modes are travelling waves, and the full forward pass of such
a model, nonlinear readout included, admits an exact operator
description~\cite{Shikder2026oscillator}. Diagonal SSMs parametrize these
numbers directly, and later selective models keep the same diagonal
structure~\cite{gu_s4,gu_s4d,orvieto_lru,gu_mamba,mamba3}: their hidden
state is a set of independent numbers, complex in S4, S4D, LRU and
Mamba-3, and real in S4D-Real, Mamba-1 and Mamba-2.

Modern selective SSMs make the transition matrix depend on the input, so that the hidden state is multiplied by a product
$\bar{A}(u_T)\cdots\bar{A}(u_1)$ of different matrices, and whether it retains the order of the inputs depends on whether these matrices commute. When the inputs describe operations, they generally do not: turning a die forward and then left leaves it in a different position than turning it left and then forward. Yet selective SSMs keep one number per channel, and at each step multiply it by a number the input chooses. Since real and complex multiplication commute, reordering the steps leaves the state unchanged, and no readout can recover the order. The limitation lies in the number system, not in the recurrence or the training.

Tasks of this nature, where the input consists of a sequence of operations and the
output is their composition, are referred to as \emph{state tracking}. Examples
include determining which face of a die appears on top, how a sequence of swaps
rearranges a set of objects, and the values that variables assume as a program executes. State tracking serves as a stringent test of whether a model has learned an underlying rule rather than relying on shortcuts, since a rule remains valid for sequences that are much longer than those encountered during training. Both transformers with fixed depth and state space models (SSMs) fail on challenging cases~\cite{merrill_illusion,grazzi,sarrof}, and stacking additional layers
recovers some, but not all, of the lost order~\cite{shakerinava,heo}.
Recent work replaces commuting numbers with more expressive transitions, such as dense unitary matrices~\cite{unitary_rnn}, products of Householder
reflections~\cite{grazzi,deltaproduct}, and permutation matrices~\cite{pdssm}. Theoretical analyses accompanying these advances each focus on a specific
family of transitions, such as complex diagonal layers~\cite{shakerinava},
abelian stacks analyzed using Lie algebras~\cite{heo}, and, in concurrent
work, rotations in $SU(2)$, $SO(3)$ and $SO(N)$, for which a single-layer
criterion, group laws that exclude $S_5$, and two-layer cascades that
achieve it are derived~\cite{jia_ncssm}. What remains unestablished is which type of memory a given task requires before a family is selected. The question matters more as SSMs spread, both as the backbone of language models~\cite{gu_mamba,dao_mamba2,jamba,griffin,nemotronh,falconh1} and in scientific machine learning~\cite{hu_ssm_operator,zheng_mambano}, where many
systems evolve by composing operations that do not commute.

We now present a theoretical framework that answers this question for any linear recurrent memory. We first observe that the hidden states of such a network form a
\textit{representation}, in the algebraic sense, of the sequences it processes
(which form a monoid under concatenation): each input symbol transforms the hidden state, and a sequence applies these transformations in order. Using this correspondence, we characterize the groups of operations a single channel can track on any associative number system  (Theorem~\ref{thm:main}). The readout plays an active role in this characterization, because it determines which part of the state is meaningful, and the part it may ignore is what allows small memories to work; for heads in $SO(3)$ and $SU(2)$ it agrees with the criterion reached independently in concurrent work~\cite{jia_ncssm}. We then show that every such memory separates into independent blocks (Proposition~\ref{prop:reduction}). The idea is familiar from diagonal SSMs. When all transitions commute, they can be diagonalized together, and the memory splits into independent one-dimensional channels, each multiplied by a real or complex number; this is the Fourier decomposition that underlies diagonal SSMs. When transitions do not commute, they cannot be diagonalized, but they can still be block-diagonalized, and we show that each block is a small matrix over one of only three number systems: the real numbers, the complex numbers or the quaternions. This has two consequences. First, the transition families studied so far, such as unitary matrices, Householder products, permutations and rotations, are not separate designs but blocks of these three kinds, so the choice of memory reduces to choosing a number system and a block size for each channel. Second, order has a minimum cost: a channel whose transitions span fewer than four real dimensions commutes,
and so tracks only operations whose order does not matter
(Corollary~\ref{cor:four}). The cheapest channels that keep order are the $2\times2$ real matrices and the quaternions, each with four real dimensions.

This correspondence also allows us to infer the symmetry of the trained transition matrices of the model, enabling a form of symmetry discovery in data. Methods for discovering symmetry usually learn transformations that leave the data invariant~\cite{benton,yang_gan}.  Here, the group is instead the set of operations that the inputs encode, and because the transition matrices must contain it, it can be read directly from a trained model.

Guided by this framework, we introduce HyperSSM, in which each channel has
its own number system, enabling it to state track, without incurring additional computational cost for linear-time inference or parallel training (Sec.~\ref{sec:hyperssm}). On formal-language and group benchmarks from recent work~\cite{walker_slice,pdssm}, HyperSSM achieves exactness with one to five states, including the case of permutations of five objects, $S_5$, with 32 generators, a setting in which the best previous linear recurrent models perform at chance level. The model is trained solely on the final answer; no information about a task's group is provided, and it maintains exactness on sequences 25 times longer than those used in training. We finally apply the same memory to three physical systems in which the state is represented as an ordered product of rotations: a moving body measured by a real gyroscope, a qubit driven by a sequence of gates, and a protein backbone, in which each residue is rotated relative to the last by its torsion angles (Sec.~\ref{sec:results}). In all three cases, a single quaternion channel matches or surpasses the physical model over sequences 16 to 60 times longer than those encountered during training, whereas commutative and recurrent baselines fail. Furthermore, the learned weights recover the calibration of the sensor, the actual rotations of the gates, and the geometry of the backbone. This framework therefore provides a precise account of what linear recurrent memory can and cannot retain regarding order, while also suggesting principled approaches for designing the memory of future sequence models.


\section{Theory}

\subsection{Groups and Representations}

A \textit{group} is a set of operations that can be composed and undone. The
24 rotations that map a cube onto itself form a group: composing any two of these rotations yields another rotation in the group,
and each rotation can be inverted. A group is \textit{abelian} if the
order of composition is always commutative, as in the case of adding hours on a clock, and
non-abelian otherwise, as for the rotations of a die. The permutations of
five objects form the group $S_5$, which is a standard challenging case for state
tracking, as its word problem is complete for the circuit class
$\mathrm{NC}^1$~\cite{merrill_illusion,barrington}. In contrast, input sequences
do not form a group: they can be concatenated, with one sequence followed by another,
but cannot be inverted. Therefore, they form a \textit{monoid}, with the empty
sequence serving as the identity element.

Throughout this paper, representation is used in the algebraic sense. A \textit{representation} assigns to each element of a group or monoid an invertible element from a number system, such as a matrix, ensuring that composition corresponds to multiplication. For example, rotations of a clock hand are represented by unit complex
numbers $e^{i\theta}$, while rotations in three-dimensional space are represented by $3\times3$ rotation
matrices. Two groups can be related by a \textit{quotient}, which identifies elements
that are considered equivalent: for example, interpreting a 24-hour clock as a 12-hour
clock identifies 3:00 and 15:00. The larger group then serves as a \textit{cover} of the
smaller one, retaining one additional bit of information that the quotient omits. Finally,
a group is \textit{compact} if its elements remain bounded, as is the case for rotations;
compact transitions neither diverge nor vanish over long sequences.

\subsection{Memory as a representation}
A diagonal SSM channel updates its hidden state as
$x_k = a_k x_{k-1} + b_k$, with input term $b_k=\bar B_k u_k$, and steps combine by
\begin{equation}
(a_2,b_2)\circ(a_1,b_1) = (a_2a_1,\;a_2b_1+b_2),
\label{eq:combine}
\end{equation}
which is what makes a parallel scan possible~\cite{blelloch,martin_cundy}.
The scan needs multiplication to be associative, \emph{not} commutative.
So a channel may live in any associative number system $A$ (a real
algebra), with invertible elements $A^\times$:
\begin{equation}
x_k = a(u_k)\,x_{k-1} + \bar B(u_k)\,u_k, \qquad a(u_k)\in A^\times,
\label{eq:channel}
\end{equation}
$A=\CC$ recovers S4D, LRU and Mamba-3; positive reals recover Mamba-1 and
Mamba-2 (SM~\ref{sm:linear}); $\bar B(u_k)$ maps the input into $A$. In the
\emph{tracking regime} ($\bar B=0$, no decay, $x_0=1$; justified in
SM~\ref{sm:regime}), a word $w=s_1\cdots s_T$
leaves the state
\begin{equation}
Q(w) = a(s_T)\cdots a(s_2)\,a(s_1).
\label{eq:product}
\end{equation}
Concatenating words multiplies states: the memory is a representation of
its input. If $A$ commutes, $Q(w)$ only counts symbols; if not, it
remembers their order [Fig.~\ref{fig:intuition}(b--d)].

\subsection{Which groups can a layer track?}
Each input symbol $s$ stands for an operation $g(s)$ in a group $G$. The
\emph{target} of a sequence $w=s_1\cdots s_T$ is the composed operation
\[
\pi(w)=g(s_T)\cdots g(s_1),
\]
with $\pi$ of the empty sequence equal to the identity $e$. The model must
output $\pi(w)$, so it only needs the elements of $G$ that occur as targets.
Let $K$ be the closure of the set of targets; it is the group the model must
track. If $G$ is finite, $K$ is the subgroup generated by the $g(s)$. For
example, if a die is only ever turned about one axis, $K$ is the cyclic
group of four quarter turns, not the full rotation group of the die. If the
inputs are continuous, as for a gyroscope, $K$ is infinite. We assume
throughout that $K$ is compact, which holds for every finite group and for
groups of rotations.

\begin{definition}\label{def:tracking}
A channel \emph{tracks} $K$ if there are a continuous readout
$\rho\colon A\to\RR^d$ and a continuous injective map $\iota\colon K\to\RR^d$
such that $\rho(Q(w)) = \iota(\pi(w))$ for all words $w$ of every length.
A stack of layers tracks $K$ if the same holds with $\rho$ a continuous
readout of the states of all its layers.
\end{definition}

``Every length'' separates learning the rule from fitting the training set;
continuity of the readout rules out memorization (SM~\ref{sm:tracking}).

\begin{theorem}\label{thm:main}
Consider channels whose reachable states lie in a compact subset of
$A^\times$. (i) If a channel tracks $K$, the closure $C$ of its reachable
states is a compact subgroup of $A^\times$, and $K\cong C/N$ for a closed
normal subgroup $N$. (ii) Conversely, if $K\cong C/N$ for a compact subgroup
$C\subset A^\times$ and closed normal $N\trianglelefteq C$, there is a
channel over $A$ with transitions in $C$ that tracks $K$.
(iii) If $N$ consists of scalar multiples of $1$, as $N=\{\pm1\}$ does,
then $C$ is a central extension of $K$, a cover when $N$ is finite, and $K$
is represented \emph{projectively}.
\end{theorem}

In words: \emph{a layer tracks a group exactly when its states hold a copy
of that group, except for a component $N$ that the readout disregards.} More precisely, the states constitute a group $C$. The mapping from a state to its encoded target is a homomorphism from $C$ onto $K$, where $N$ denotes its kernel, representing the states that the readout cannot distinguish from the identity element. If $N$ is trivial, then $C$ is an exact copy of $K$ within $A^\times$. Allowing a nontrivial $N$ is significant because $C$ may then be embedded in a smaller algebra than any exact copy of $K$ would permit.

Spatial rotations illustrate this property. The group $SO(3)$ is not a subgroup of the unit quaternions, as the only quaternion of order two is $-1$, whereas $SO(3)$
contains a half-turn about every axis. Therefore, an exact representation requires a larger algebra, such as $M_3(\RR)$, which consists of nine elements. Instead, the unit quaternions $C=SU(2)$ map onto $SO(3)$ by associating $q$ with the rotation
$v\mapsto qv\bar q$. The kernel of this mapping is $N=\{\pm1\}$, since $q$
and $-q$ yield the same rotation. A quaternion channel, which utilizes four components, therefore can represent $SO(3)$, provided its readout assigns the same value to $q$ and $-q$, as in $\rho(q)=L(qq^{\mathsf T})$, where $L$ denotes a learned linear map. [Fig.~\ref{fig:intuition}(e)]. The same reasoning applies to every finite group of rotations. In particular, the rotation group of the icosahedron, $A_5$, does not have an exact representation in $\HH$, but its preimage, the binary icosahedral group $2I$, does. Therefore, a single quaternion channel can represent $A_5$. Theorem~\ref{thm:main} specifies when such a readout is necessary and what information it may ignore. The proof is provided in SM~\ref{sm:tracking}.

\subsection{Every memory splits into $\RR$, $\CC$ and $\HH$ blocks}

\begin{proposition}[Semisimple reduction]\label{prop:reduction}
If a channel over $A$ tracks $K$ with reachable states in a compact subset
of $A^\times$, let $C$ be as in Theorem~\ref{thm:main} and
$B=\mathrm{span}_\RR C$. Then $B$ is a semisimple subalgebra of $A$, the
same channel regarded as a channel over $B$ tracks $K$, and
\begin{equation}
B\;\cong\;\prod_i M_{n_i}(D_i),\qquad D_i\in\{\RR,\CC,\HH\}.
\label{eq:wedderburn}
\end{equation}
The factors evolve independently and meet only in the readout; in a
suitable basis their transitions are orthogonal, unitary or
quaternion-unitary.
\end{proposition}

This represents the order-preserving analog of the Fourier decomposition:
A time-invariant memory decomposes into complex clock hands, while a general memory decomposes into rotating matrix blocks over $\RR$, $\CC$, or $\HH$~\cite{broecker,lam}. These are the only number systems worth considering (proof in SM~\ref{sm:reduction}).

\begin{corollary}\label{cor:four}
A channel over an algebra of dimension less than four tracks only abelian
groups. In dimension four, the algebras that track non-abelian groups are
the $2\times2$ real matrices $M_2(\RR)$ and the quaternions $\HH$.
\end{corollary}

To preserve order, a channel's transitions need at least four real
dimensions. This bound is on the transitions, not on the state. For example, in the recurrence $x_k=a(u_k)\,x_{k-1}$, the transition and the state can differ in size. $a(u_k)$ is an element of the algebra, whereas the state $x_k$ only needs to be a vector upon which it acts. For instance, over $M_2(\RR)$, each transition is represented by a $2\times2$ matrix, consisting of four real numbers, while the state can be a vector in $\RR^2$, which consists of two real numbers. This configuration is sufficient to track the symmetries of a polygon
(SM~\ref{sm:reduction}). In contrast, over $\HH$ there is no such reduction: the state is a quaternion, which consists of four real numbers, the same size as the transition.

\begin{figure}[tb]
\centering
\includegraphics[width=\columnwidth]{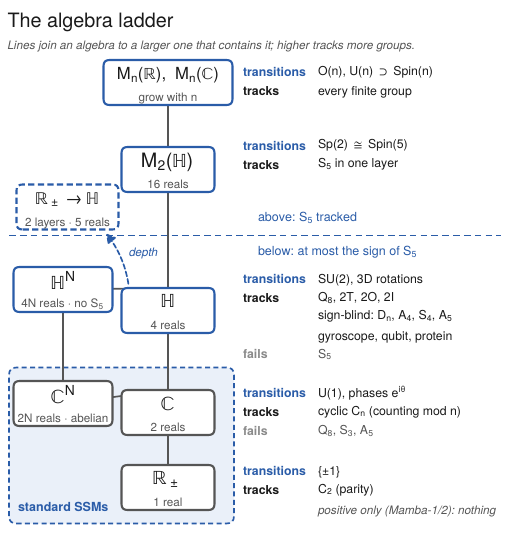}
\caption{\textbf{The algebra ladder.}  Each box represents an algebra with its associated properties. The number of real components is indicated; on the right, bounded transitions and the groups tracked by a single channel are shown (Theorem~\ref{thm:main}). For fewer than four components, every algebra is commutative (Corollary~\ref{cor:four}). Increasing the width of a layer results in lateral movement and does not cross the dashed line, above which $S_5$ is tracked. In contrast, increasing depth does: a sign channel beneath a quaternion channel tracks $S_5$ with five numbers (Proposition~\ref{prop:depth}; Table~\ref{tab:groups}). Shaded regions indicate channels of standard diagonal SSMs.
}
\label{fig:ladder}
\end{figure}

\subsection{The algebra ladder}
Theorem~\ref{thm:main} and
Proposition~\ref{prop:reduction} turn ``can this layer learn this task?''
into a lookup (Fig.~\ref{fig:ladder}; details in SM~\ref{sm:ladder}):
\begin{itemize}
\item \emph{One real number:} parity only~\cite{grazzi}.
\item \emph{Complex numbers, any width:} counting modulo $n$; only abelian groups.
\item \emph{One quaternion (4 numbers):} $Q_8$, $A_4$, the die $S_4$, and the
icosahedral rotations $A_5$.
\item \emph{$2\times2$ quaternion matrix (16 numbers):} $S_5$.
\item \emph{A sign beneath a quaternion (5 numbers, two layers):} $S_5$, by depth (Proposition~\ref{prop:depth}).
\end{itemize}

\subsection{Depth helps, but only up to solvable groups}
The stacking of commutative layers restores a degree of order. For the six symmetries of a triangle, one layer monitors the parity of reflections, while a subsequent complex layer rotates by $\pm120^\circ$ according to that parity.

\begin{proposition}[Depth]\label{prop:depth}
In the tracking regime, consider a stack of $L$ layers in which every layer
is commutative with transitions in a compact group, the transitions of each
layer are functions of the input symbol and of the previous states of the
layers below, and every layer below the top has finitely many reachable
states. Such a stack tracks a finite group $K$ if and only if $K$ is
solvable of derived length at most $L$.
\end{proposition}

So complex stacks reach $Q_8$, $A_4$ and the die, but no such stack of
commutative layers reaches $A_5$ or $S_5$. In contrast, a single quaternion channel
tracks $A_5$. The finiteness assumption is satisfied whenever the lower layers
track finite groups with exact transitions, as demonstrated in the construction and in our snapped models. Without this assumption, the closure argument of Theorem~\ref{thm:main} does not apply directly (SM~\ref{sm:depth}). For complex diagonal layers, this result corresponds to the theorem of Shakerinava et al.~\cite{shakerinava}, which is proved at finite precision (see also~\cite{heo}, where tracking up to a bounded length is achieved with depth logarithmic in that length). This result serves as the finiteness assumption in our context; the
representation perspective provides this property for any commutative channel algebra and situates it alongside Theorem~\ref{thm:main}. In contrast to circuit-complexity bounds~\cite{merrill_illusion,sarrof},
this bound is unconditional within its model (see proof in SM~\ref{sm:depth}).


\begin{table*}[t]
\caption{\label{tab:lang}\textbf{Regular languages.} Accuracy (\%) on lengths 40--256 after training on 3--40, mean $\pm$ s.d.\ over 5 runs. Baselines
from~\cite{pdssm}. HyperSSM algebras, with numbers of state: $\RR_\pm$ (1),
$\CC$ (2), $M_{16}(\CC)$ (32). Discovered $n$: the smallest $n$ for which
snapping each complex transition to an $n$-th root of unity is exact on
validation, found after training without being told the task. Details in
SM~\ref{sm:bench}.}
\centering
\includegraphics[width=\textwidth]{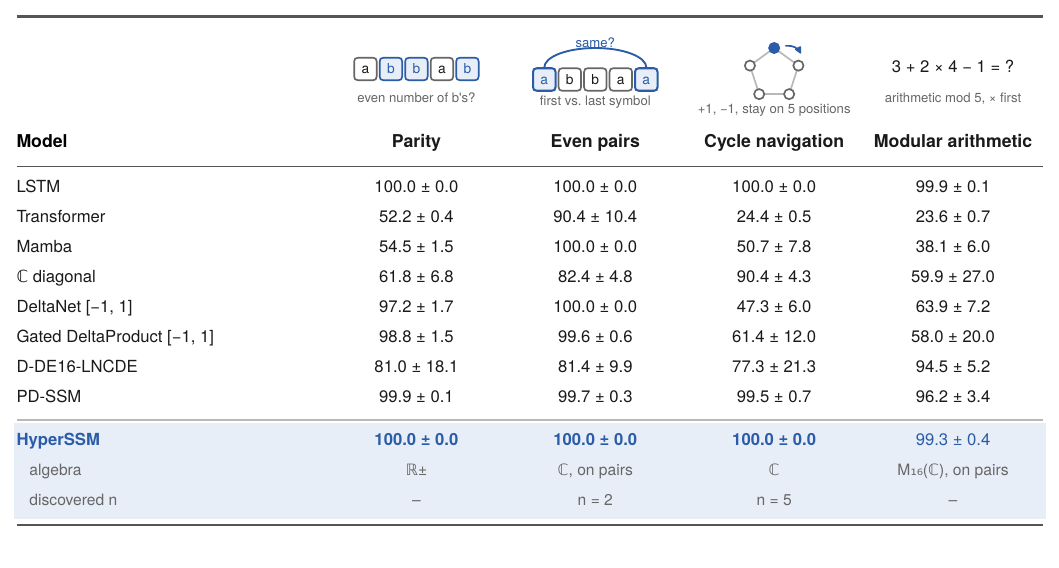}
\end{table*}

\begin{table*}[t]
\caption{\label{tab:groups}\textbf{Group word problems.} Accuracy (\%) on
lengths 40--256 and, for HyperSSM, 256--1000, after training on 3--40; best
of 3 runs. Chance: 1.67\% ($A_5$), 0.83\% ($S_5$); --: not reported.
Baselines from~\cite{pdssm} (PD-SSM: 128 complex numbers of state).
HyperSSM: $\HH$ (4), $\RR_\pm\!\to\!\HH$ (5). All runs in SM~\ref{sm:bench}.}
\centering
\includegraphics[width=\textwidth]{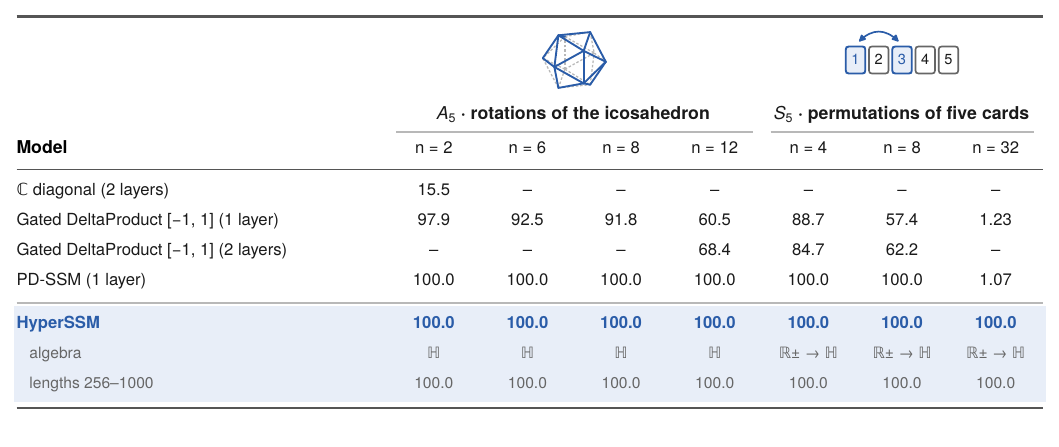}
\end{table*}

\section{HyperSSM}\label{sec:hyperssm}
\begin{figure}[tb]
\centering
\includegraphics[width=\columnwidth]{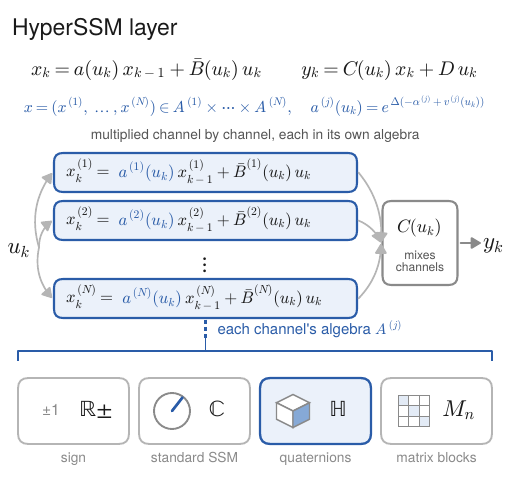}
\caption{\textbf{A HyperSSM layer.} Each channel follows the linear
recurrence of a diagonal SSM, with input $u_k$, hidden state $x^{(j)}_k$,
transition $a^{(j)}(u_k)$ and input term $\bar B^{(j)}(u_k)u_k$, with
multiplication in its own algebra $A^{(j)}$. Channels evolve independently,
mix only in the readout $C$, and are computed by the same parallel scan.}
\label{fig:hyperssm}
\end{figure}

The theory says which number system a task needs;
HyperSSM lets each channel choose it, and changes nothing else
(Fig.~\ref{fig:hyperssm}).

\begin{definition}\label{def:hyperssm}
A HyperSSM layer with input $u_k\in\RR^d$ and channels over algebras
$A^{(1)},\dots,A^{(N)}$ has hidden state
$x_k=(x^{(1)}_k,\dots,x^{(N)}_k)\in A^{(1)}\times\cdots\times A^{(N)}$, updated
channel by channel as
\begin{equation}
\begin{aligned}
x^{(j)}_k &= a^{(j)}(u_k)\,x^{(j)}_{k-1} + \bar B^{(j)}(u_k)\,u_k,\\
a^{(j)}(u_k) &= e^{\Delta\left(-\alpha^{(j)} + v^{(j)}(u_k)\right)},\\
y_k &= C(u_k)\,x_k + D\,u_k,
\end{aligned}
\label{eq:hyperssm}
\end{equation}
where the generator $v^{(j)}(u_k)\in A^{(j)}$ and the input matrix
$\bar B^{(j)}(u_k)\colon\RR^d\to A^{(j)}$ are linear in the input,
$\alpha^{(j)}\ge0$ is a learned decay rate, $\Delta>0$ a step size, and the
exponential is the power series in $A^{(j)}$. The readout $C(u_k)$ acts on
all channels, on $x_k$ or on a sign-invariant function of it such as
$x^{(j)}_k x^{(j)\mathsf T}_k$, and $D$ is a skip connection.
\end{definition}

This is the selective SSM $x_k=\bar A(u_k)x_{k-1}+\bar B(u_k)u_k$,
$y_k=C(u_k)x_k+Du_k$ of Mamba~\cite{gu_mamba} with $\bar A(u_k)$ block-diagonal,
one block per channel, multiplying in that channel's algebra. With complex
channels this \emph{is} a diagonal SSM. With a quaternion
channel, the input chooses a rotation axis as well as an angle; that is the
whole difference. Real channels use a learned sign, $a=\pm1$. All channels
share the scan of Eq.~\eqref{eq:combine}, so HyperSSM keeps $O(T)$ work and
$O(\log T)$ depth (SM~\ref{sm:hyperssm}) (where $T$ is the sequence length). The input enters both through the
transition and, as in any SSM, through $\bar B$; all experiments here are in
the tracking regime, $\bar B=0$, where it acts through the transition alone.

\section{Results}\label{sec:results}

\subsection{Exact state tracking on benchmarks}
We use the formal-language and group benchmarks of~\cite{walker_slice,pdssm}: train in lengths 3--40, test on 40--256, and for groups also on 256--1000. 
For each task, the theory selects the minimal required representation: one real channel for parity, one complex channel for cycle navigation and even pairs,
one quaternion for $A_5$, and a sign channel combined with a quaternion for $S_5$.
No information about the task's group is provided to the model.
(Protocol details are in SM~\ref{sm:bench}.) Real channels are constrained to exactly $\pm1$ throughout. Complex channels are trained continuously and then discretized to the $n$-th roots of unity, where $n$ is the smallest value that achieves exact validation performance and is determined from the trained model (as described in Theorem~\ref{thm:main}). Quaternion channels are never
snapped.

HyperSSM is exact on parity, even pairs and cycle navigation with one or two
numbers of state (Table~\ref{tab:lang}), and on every $A_5$ and $S_5$ task
with four and five numbers of state: one quaternion for $A_5$, and a sign
channel beneath a quaternion for $S_5$ (Table~\ref{tab:groups}). For $S_5$ with 32 generators, PD-SSM (128 complex numbers) achieves 1.07\% accuracy, and Gated DeltaProduct achieves 1.23\%, which is near chance level (0.83\%), while HyperSSM remains exact. HyperSSM maintains exact performance up to 25 times the training sequence length. For modular arithmetic, a dense $M_{16}(\CC)$ block achieves $99.3\pm0.4\%$ accuracy; although an exact memory solution for three numbers exists, it is not discovered through training (SM~\ref{sm:modarith}).

\begin{figure*}[t]
\centering
\includegraphics[width=\textwidth]{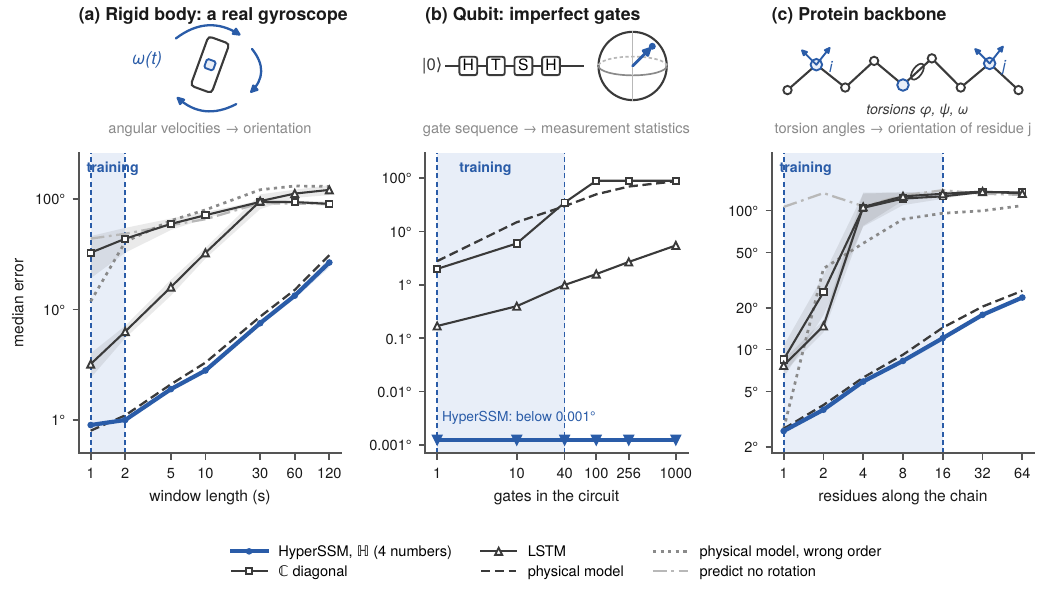}
\caption{\textbf{Applications: Physical systems in which the state is represented as an ordered product of rotations.} Median error as a function of sequence length; training lengths are shaded; mean $\pm$ standard deviation over three random seeds. (a) Orientation of a moving body measured by a real gyroscope (BROAD; held-out recordings); physical model: integration of the sensor readings. (b) Bloch vector of a qubit following a gate sequence, with imperfect gates (2\% under-rotation and $2^\circ$ axis tilt); physical model: simulation using ideal gates; HyperSSM achieves errors below the lower axis limit
($10^{-3}$ degrees) at all sequence lengths; complex diagonal SSM uses three layers. Details on Clifford gates and ideal Clifford+$T$ gates are provided in SM~\ref{sm:qubit}. (c) Orientation of residue $j$ relative to the frame of residue $i$, computed from the torsion angles, on held-out PDB proteins; physical model: reconstruction using standard bond geometry. In all three cases, HyperSSM uses a single quaternion channel (four state variables); the complex diagonal SSM uses 128 state variables.}
\label{fig:apps}
\end{figure*}

\subsection{Applications}
The composition rule HyperSSM
is built on is not specific to benchmarks. Whenever the state of a physical
system is an ordered product of rotations, a quaternion channel is an exact
memory for it, and a commutative one cannot be (Corollary~\ref{cor:four}).
We test three such systems with the same architecture and protocol: train
on short sequences, test on long ones, and compare with a complex diagonal
SSM, an LSTM, and the physical model itself.

\subsubsection{Rigid bodies: a real gyroscope}
A gyroscope measures angular velocity, and orientation is determined as the ordered product of these incremental rotations. Reversing the order of the same sequence of small rotations yields a different orientation and leads to a $41^\circ$ error after 2 seconds. On the BROAD dataset~\cite{broad}, models process gyroscope windows to predict rotation, with training performed on 1 to 2 second intervals and testing conducted for durations up to 2 minutes (experimental setup described in SM~\ref{sm:imu}). HyperSSM utilizes a single quaternion channel with 12 parameters. A complex diagonal state space model (SSM) with 128 state variables does not outperform a baseline that predicts no rotation, as anticipated by Corollary~\ref{cor:four}. An LSTM accurately fits the training intervals but fails to generalize to longer durations. HyperSSM matches gyroscope integration performance and exhibits reduced drift over extended time horizons ($26.5^\circ$ compared to $30.9^\circ$ at 2 minutes), due to its ability to learn the sensor's calibration [Fig.~\ref{fig:apps}(a)].

\subsubsection{Qubits: simulating and characterizing gates}
Single-qubit gates correspond to unit quaternions, implying that a quaternion channel represents a qubit. The Born rule is quadratic in the quantum state; therefore, measurement statistics are precisely determined by this relationship. This results in a sign-invariant readout $L(qq^{\mathsf T})$, while the unobservable global phase $-1$ corresponds to the component $N$ in Theorem~\ref{thm:main} that the readout must
ignore. The models process a random gate sequence and predict the Bloch vector of
the final state. The Born rule is then used to determine the probability of each measurement outcome; the models are trained on circuits containing 1 to 40 gates and evaluated on circuits with up to 1000 gates (setup in SM~\ref{sm:qubit}). The Clifford gates $\{H,S\}$ operate on the Bloch sphere as
the 24 rotations of a cube, which form a solvable group. Including the $T$ gate renders the set of gates dense in $SU(2)$. HyperSSM achieves exact results in both cases: its Bloch vectors remain within
$10^{-3}$ degrees of the true values for all sequence lengths up to 1000 gates, which is 25 times the training length. The learned gates correspond to $H$, $S$, and $T$ to three decimal places in both angle and axis [Fig.~\ref{fig:apps}(b) and SM~\ref{sm:qubit}]. A three-layer
complex diagonal stack, which is sufficiently deep to represent the Clifford group by Proposition~\ref{prop:depth}, does not identify this structure during training and remains near $90^\circ$ deviation from about 100 gates onward. An LSTM can model the finite Clifford group
to within approximately $0.2^\circ$, but exhibits drift on Clifford+$T$, reaching about $6^\circ$ at 1000 gates. In hardware implementations, gates are imperfect. When each gate under-rotates by 2\% about an axis tilted by $2^\circ$, a simulator employing ideal gates drifts to nearly $90^\circ$ at 1000 gates, whereas HyperSSM remains exact. From measurement data alone, HyperSSM recovers the actual gates to three decimal places ($H$: $176.400^\circ$ learned and true; $S$: $88.200^\circ$; $T$: $44.100^\circ$; angle between the $H$ and $S$ axes: $46.643^\circ$), representing a learned form of gate set characterization~\cite{nielsen_gst}.

\subsubsection{Proteins: the backbone as a chain of rotations}
The orientation of residue $j$ relative to the frame of residue $i$ is represented as the ordered product of the rotations between consecutive residues. Each rotation is determined by three torsion angles $(\psi,\omega,\phi)$ and nearly constant bond angles, and the residue frames correspond to those constructed in structure prediction~\cite{alphafold}. A sequence of $n$ residues is represented as $6n+1$ tokens, where each token corresponds to a torsion angle with its
associated angle or a bond. HyperSSM operates as a single quaternion channel within the tracking regime with generator $v_t=W(e_{\tau_t}\otimes[1,\alpha_t])$, 48 parameters as detailed in SM~\ref{sm:protein}. The evaluation uses 554 high-resolution crystal structures from the Protein Data Bank~\cite{pdb}, partitioned by entry. The models are trained on sequences of 1 to 16 residues and evaluated on held-out proteins containing up to 256 residues. Composing the same per-residue rotations in an incorrect order results in $38^\circ$ of error after two residues. The complex diagonal SSM and the LSTM exceed $100^\circ$ of error from four residues onward, within the training range. HyperSSM is more accurate than reconstruction using standard bond geometry~\cite{engh_huber} across all sequence lengths from 2 to 64 residues ($12.1^\circ$ versus $14.4^\circ$ at 16,
$23.7^\circ$ versus $26.5^\circ$ at 64; Fig.~\ref{fig:apps}(c)), as it
captures the average geometry of actual protein backbones. The remaining error is attributable to the variation in bond angles between residues, which cannot be addressed by any model based on torsion angles alone.

\subsection{Reading the group from the transition}

Theorem 1 fixes what a trained channel must contain: a copy of the task's group structure. The group can therefore be read from the learned transitions and compared with the true one.

\begin{proposition}[Certificate and horizon]\label{prop:read}
In the tracking regime: (i) If the states of a model take finitely many
values, as when its transitions generate a finite group $C$, the model
tracks $K$ at every length if and only if its readout is correct on every
pair $(Q(w),\pi(w))$ reachable in $C\times K$; these are at most
$|C|\,|K|$ pairs, found by breadth-first search. (ii) If a channel's
transitions have norm at most one and lie within $\varepsilon$ of those of a
model certified by (i), and the readout's decision regions contain the balls
of radius $r$ around that model's states, the channel is exact on every word
of length $T<r/\varepsilon$.
\end{proposition}

Part (i) reformulates the condition of being “exact at every length” into a finite verification; Part (ii) determines, based solely on the weights, the extent to which a continuous model is guaranteed to generalize (proof in SM~\ref{sm:interp}). The decoding procedure is explicit. A
quaternion transition corresponds to a rotation by $2\arccos|a_0|$ about the axis
$(a_1,a_2,a_3)$. Consequently, each $A_5$ generator must rotate by $180^\circ$,
$120^\circ$, $72^\circ$, or $144^\circ$ about a symmetry axis of an
icosahedron, and the learned sign of each $S_5$ generator must correspond to the parity of its permutation. The discretized complex
channels already demonstrated this mechanism: training identified $n=2$ for even pairs and $n=5$ for cycle navigation without explicit instruction. In the gyroscope model, the 12 parameters represent the sensor's scale, axis misalignment, and bias; in the qubit model, the transitions correspond to the gates' actual rotations; in the protein model, they represent the torsion and bond rotations of the backbone.

\section{Discussion}
Convolutional networks taught us that a symmetry of the
data should shape the model. Early SSMs had time-shift symmetry, and complex
numbers were its natural basis. Selective models broke that symmetry; what
governs them is the group of the operations they read, and the right basis
is a set of blocks over $\RR$, $\CC$ and $\HH$. Real and complex numbers are
the abelian case, and stacks of them the solvable one. Because a
tracking memory must be a representation, the same theory says where to
look inside a trained model and what to expect there.

\subsection{Scientific machine learning}
Many systems in the sciences
evolve through the composition of non-commuting operations. For these systems, the
algebraic structure of memory constitutes a physical property rather than a modeling assumption. Examples include rigid-body motion, spin and qubit dynamics, polarization optics, parallel transport along paths in gauge theory, and the braiding of anyons. These phenomena can be described as ordered products in $SU(2)$, $SO(3)$, and related groups (SM~\ref{sm:outlook}).
For each case, Corollary~\ref{cor:four} excludes the possibility of a commutative memory, and a quaternion or matrix channel provides an exact representation by construction. The qubit experiment demonstrates a practical application in quantum hardware: a sequence model in which the transitions correspond to gates learned from measurements, characterizes the gates in a manner analogous to gate set tomography~\cite{nielsen_gst}, and also predicts the behaviour of circuits longer than those used during training. When decay is present ($\alpha>0$), the
same channel can account for decoherence.

\subsection{Protein structure}
Structure prediction and design represent
a protein as a sequence of rigid residue frames and compose these frames, as implemented in the
structure module of AlphaFold~\cite{alphafold} and in diffusion models that
generate protein backbones directly using rigid motions~\cite{yim_framediff,rfdiffusion}.
The backbone experiment demonstrates the simplest form of this compositional approach, carried
out by a linear memory with a parallel scan. This method is exact in its algebraic formulation, linear in
sequence length, and interpretable, since each learned rotation corresponds to a torsion angle or a bond.
Channels utilizing dual quaternions, an associative algebra that composes
rotations and translations together, could encode both positions and
orientations. However, the group of rigid motions associated with dual quaternions is not compact, so this approach lies
beyond the scope of the present theorems but remains within the same recurrence framework.

\subsection{Symmetry discovery}
Methods for discovering symmetry learn
which transformations leave the data invariant~\cite{benton,yang_gan}. A
tracking memory finds a different symmetry: the group generated by the
operations it processes. Theorem~\ref{thm:main} states that a model capable of tracking must contain this group, enabling direct extraction rather than requiring a search. During training, $\ZZ_2$ and $\ZZ_5$ were identified in the complex channels, the actual rotations of imperfect gates in the qubit model, and the torsion geometry of the backbone in the protein model. A general procedure, from
data to the group and its representation, constitutes a logical next step.

\subsection{Long context}
These differences matter most over long sequences. A memory in the right algebra composes exactly, so its errors do not compound: on the benchmarks it stays exact at 25 times the training length, it matches the physical model at 60 times on the gyroscope and at 16 times along the protein chain, and Proposition~\ref{prop:read}(ii) bounds how long a continuous model stays exact, $T<r/\varepsilon$. A memory in the wrong algebra fails beyond the training range at any width. As language models are asked to follow long programs, agent trajectories and documents in which
order matters, the number system of their recurrent memory becomes a design
variable alongside width and depth. Quaternion and matrix channels add a
constant factor to the scan and could sit beside complex channels in the
hybrid recurrent and attention models now in use~\cite{griffin,jamba};
whether they improve such models at scale is the main open question.

\subsection{Relation to prior and concurrent work}
Earlier analyses each begin by fixing a family of transitions and ask what
that family can do: complex diagonal layers~\cite{shakerinava} and abelian
stacks through Lie-algebra extensions~\cite{heo}. The present work reverses
the question and asks which memory a task requires. This gives three results
that a family-by-family analysis does not. First, Theorem~\ref{thm:main}
holds for a channel over any associative algebra, so fixed families become
special cases: the depth theorem of~\cite{shakerinava} is
Proposition~\ref{prop:depth} for complex channels, and for rotation heads
the single-layer criterion, the exclusion of $S_5$ and the two-layer cascades
given in concurrent work~\cite{jia_ncssm} correspond to
Theorem~\ref{thm:main}, Lemma~\ref{lem:su2} and Proposition~\ref{prop:depth}
for $\HH$. Second,
Proposition~\ref{prop:reduction} shows that the search over families is
finite: every memory is a sum of blocks over $\RR$, $\CC$ and $\HH$, and the
smallest memory that tracks a task can be read off its group. Third, because
the memory must be a representation, a trained model is a checkable object:
Proposition~\ref{prop:read} certifies exactness at every length from the
weights and bounds how long a nearly exact model stays exact.

The same starting point changes what the experiments can show. Because the
theory fixes the memory before training, HyperSSM can be trained under the
strictest protocol and still be exact. On $S_5$ with 32 generators, where the
best previous linear recurrent models are at chance~\cite{pdssm}, it is
learned from the final answer alone, with five numbers of state, a linear
readout and no information about the group, and it stays exact on sequences
25 times longer than those seen in training. On physical systems, a single
quaternion channel regresses orientation from a real gyroscope against motion
capture at 60 times the training length, and its weights recover the
calibration of the sensor, the actual rotations of imperfect qubit gates, and
the geometry of protein backbones.

\subsection{Open problems}
Open are a
quantitative theory under decay and noise, the depth bound for stacks whose
lower layers take a continuum of values, extensions to resets and all
regular languages~\cite{krohn_rhodes}, whether this legibility
survives in large models that mix such channels with attention, and how to
train the smallest memory when it is not the easiest to find
(SM~\ref{sm:outlook}).

\section*{Code availability}
Code to reproduce all experiments, including the HyperSSM implementation,
benchmark protocols and trained models, is available at
\url{https://github.com/anifshikder/HyperSSM}.


\bibliography{references}

\clearpage
\onecolumngrid
\begin{center}
\textbf{\large Supplemental Material}
\end{center}
\setcounter{equation}{0}
\setcounter{figure}{0}
\setcounter{table}{0}
\setcounter{section}{0}
\setcounter{secnumdepth}{3}
\renewcommand{\theequation}{S\arabic{equation}}
\renewcommand{\thefigure}{S\arabic{figure}}
\renewcommand{\thetable}{S\arabic{table}}
\renewcommand{\thesection}{S\arabic{section}}
\renewcommand{\thelemma}{S\arabic{lemma}}
\renewcommand{\thesubsection}{\thesection.\arabic{subsection}}

\section{Background on groups}\label{sm:background}

This section illustrates each algebraic concept with the die.

A \emph{group} is a set of operations that can be composed and inverted.
The 24 rotations that map a cube onto itself form a group: composing two
rotations gives another rotation, and each rotation has an inverse. The
group is \emph{abelian} if composition does not depend on order; the
rotations of the cube are not abelian, whereas addition of hours on a clock
modulo 12 is. The group of all permutations of five objects is $S_5$; it is
the standard hard case in the literature, as its word problem is complete
for NC$^1$~\cite{barrington,merrill_illusion}.

A \emph{representation} of a group assigns to each operation an invertible
element of a number system (for example a matrix) so that group composition
becomes multiplication. Rotations of 3D space are represented by $3\times3$
rotation matrices; rotations of a clock hand by unit complex numbers
$e^{i\theta}$.

A \emph{quotient} identifies elements that should be considered equivalent:
reading a 24-hour clock as a 12-hour clock merges 3:00 and 15:00. A
\emph{cover} retains extra information: the 24-hour clock covers the
12-hour clock by keeping one bit the latter omits. A group is
\emph{compact} if its elements remain bounded, as rotations do; in a memory,
compact transitions neither diverge nor vanish over long sequences.

\section{Linear memory and time-shift symmetry}\label{sm:linear}

\emph{Discretization.} An SSM layer is a discretized linear system
$\dot x = Ax + Bu$, $y = Cx$. With step size $\Delta$,
\begin{equation}
x_k = \bar A x_{k-1} + \bar B u_k,\qquad
\bar A = e^{\Delta A},\quad \bar B = A^{-1}\bigl(e^{\Delta A}-I\bigr)B.
\label{eq:ssm}
\end{equation}
Deep SSMs stack such layers with nonlinearities in between. In practice $A$
is diagonal over $\CC$~\cite{gu_s4d,orvieto_lru}, so the state splits into
scalar recurrences
\begin{equation}
x_k = a_k x_{k-1} + b_k, \qquad a_k = e^{\Delta\lambda},\quad b_k=\bar B_k u_k,
\label{eq:scalar}
\end{equation}
one per eigenvalue $\lambda$ of $A$; each is a \emph{channel}, a damped
oscillator that scales the state by $|a_k|$ and rotates it by $\arg a_k$.
Selective SSMs such as Mamba~\cite{gu_mamba,dao_mamba2} make $\Delta$, $B$
and $C$ input-dependent. Each step is an affine map $x\mapsto ax+b$, and the
composition rule Eq.~\eqref{eq:combine} is associative, so all prefix states
follow from a parallel scan~\cite{blelloch,martin_cundy} in $O(\log T)$
rounds and $O(T)$ work. Linear attention~\cite{katharopoulos} and
DeltaNet~\cite{yang_deltanet} have the same structure with matrix-valued
states.

\emph{Why complex numbers.} If transitions do not depend on the input,
shifting the input shifts the output, and the layer is a convolution. The
functions a convolution leaves invariant up to scale are the exponentials
$k\mapsto\lambda^k$, so such a memory is a bank of independent channels
evolving as $\lambda^k$; without damping, clock hands $e^{i\theta k}$. These
are the representations of the abelian group of time shifts, each needing a
single complex number: a diagonal SSM is the Fourier decomposition of a
time-invariant memory. Rotary position embeddings~\cite{su_roformer} use the
same clock hands.

\emph{Selective models keep commuting channels.} Selective SSMs break
time-shift symmetry but keep scalar channels: positive real in Mamba-1 and
Mamba-2, complex in Mamba-3. All their transitions commute. A complex channel
that turns by $\theta_1$ then $\theta_2$ ends where it would in reverse
order; its state can depend on symbol counts and, with damping, on timing,
but not on the order of non-commuting operations. This explains why
Mamba-3, whose complex update is a data-dependent rotary embedding, solves
parity and modular arithmetic~\cite{mamba3} (built from counters modulo
$n$), while the die stays out of reach.

\emph{Prior use of quaternions.} Quaternion arithmetic has been used in
neural networks, including recurrent ones~\cite{parcollet_qrnn}, to couple
the components of multidimensional features. Here quaternions instead serve
as the non-commuting transitions of a linear memory that keeps its parallel
scan.

\emph{Quaternions.} A quaternion is $q=q_0+q_1\mathbf i+q_2\mathbf
j+q_3\mathbf k$ with $\mathbf{ij}=\mathbf k$ and $\mathbf{ji}=-\mathbf k$.
Unit quaternions represent 3D rotations, which is why graphics and robotics
use them; multiplying them composes rotations, and order matters, as for the
die [Fig.~\ref{fig:intuition}(d)].

\section{Reduction to the tracking regime}\label{sm:regime}

The main text studies channels with no input term, no decay and $x_0=1$.
This section shows that none of the three restrictions loses tracking power.
Throughout, a symbol $s$ acts on the state of a channel by the affine map
$T_s(x)=a(s)\,x+b(s)$, with $b(s)=\bar B(s)u(s)$ and $u(s)$ the embedding of
$s$, and a word acts by $T_w=T_{s_T}\circ\cdots\circ T_{s_1}$. The norm of an
element of $A$ is the operator norm of left multiplication by it.

\begin{lemma}[Compact monoids]\label{lem:monoid}
A compact submonoid $S$ of a Hausdorff topological group is a subgroup.
\end{lemma}
\emph{Proof.} Let $x\in S$. The powers $x^n$ lie in $S$, so a subsequence
$x^{n_i}$ converges, and we may take $m_i=n_{i+1}-n_i\to\infty$. Then
$x^{m_i}=x^{n_{i+1}}(x^{n_i})^{-1}\to1$, so $x^{m_i-1}\to x^{-1}$; as
$m_i-1\ge0$ and $S$ is closed, $x^{-1}\in S$. $\square$

\begin{lemma}[Input term and initial state]\label{lem:affine}
Suppose the maps $\{T_w\}$ of a channel lie in a compact subset of the
affine group of $A$ and their linear parts in a compact subset of
$A^\times$. If the channel, from an initial state $x_0$, tracks $K$, then
the channel over $A$ with the same transitions $a(s)$, no input term and
$x_0=1$ tracks $K$ with a suitable readout.
\end{lemma}
\emph{Proof.} The closure $\Gamma$ of $\{T_w\}$ is a compact submonoid of
the affine group, hence a compact group by Lemma~\ref{lem:monoid}. With
Haar measure $d\gamma$, the point $x_*=\int_\Gamma\gamma(0)\,d\gamma$ is
fixed by every element of $\Gamma$, because affine maps commute with
averaging and Haar measure is invariant. In the coordinate $z=x-x_*$ every
symbol acts linearly, $z\mapsto a(s)z$, so the state after $w$ is
$x_*+Q(w)z_0$ with $z_0=x_0-x_*$ and $Q(w)$ as in Eq.~\eqref{eq:product}.
If $\rho$ is the readout of the original channel, $\rho'(Q)=\rho(x_*+Qz_0)$
is continuous on $A$ and $\rho'(Q(w))=\iota(\pi(w))$. $\square$

So the input term only moves the point about which the state turns, and
the initial state can be folded into the readout.

\begin{lemma}[Decay]\label{lem:decay}
Let $K$ be finite. Split a layer's state as $x=(x^{\mathrm n},x^{\mathrm c})$
into channels whose transitions have norm at most one and channels whose
transitions have norm at most $\lambda<1$, and suppose all reachable states are bounded.
If the layer tracks $K$, then the channels $x^{\mathrm n}$ alone track $K$.
In particular, a layer whose channels all decay tracks only the trivial
group.
\end{lemma}
\emph{Proof.} Let $R$ be the bounded set of reachable states. On its compact
closure the readout $\rho$ is uniformly continuous; let $\delta>0$ be half
the smallest distance between the finitely many points of $\iota(K)$, and
let $\eta>0$ be such that states within $\eta$ have readouts within
$\delta$. Choose a word $u$ so long that $\lambda^{|u|}\operatorname{diam}R<\eta/2$.
For words $w,w'$, the decaying parts after $u$ satisfy
$\|x^{\mathrm c}(wu)-x^{\mathrm c}(w'u)\|<\eta/2$, while the norm-one parts
satisfy $\|x^{\mathrm n}(wu)-x^{\mathrm n}(w'u)\|\le\|x^{\mathrm n}(w)-x^{\mathrm n}(w')\|$.
Hence, if $\|x^{\mathrm n}(w)-x^{\mathrm n}(w')\|<\eta/2$, the readouts of
$wu$ and $w'u$ are within $\delta$, so $\pi(u)\pi(w)=\pi(u)\pi(w')$ and
$\pi(w)=\pi(w')$. The answer is therefore a uniformly locally constant
function of $x^{\mathrm n}$ on the reachable set; it extends continuously to
the closure, and composing with $\iota$ and extending to all of the algebra
of $x^{\mathrm n}$ (Tietze) gives a readout of $x^{\mathrm n}$ alone. With
no norm-one channels, the answer is constant, so $K$ is trivial. $\square$

Decay therefore lets a channel weigh recent inputs, but every bit of order
kept at every length is kept by the channels that do not decay. Together,
Lemmas~\ref{lem:affine} and~\ref{lem:decay} reduce the analysis to the
\emph{tracking regime}: no input term, no decay, $x_0=1$, and one transition
$a(s)$ per input symbol $s$. After a word $w$ the state is
Eq.~\eqref{eq:product}, and $Q(ww')=Q(w')\,Q(w)$. Channels whose norm varies
from symbol to symbol, some at one and some below, are not covered; this is
part of the quantitative theory under decay listed as open.

\section{Tracking: definition and proof of Theorem~\ref{thm:main}}\label{sm:tracking}

\emph{On the definition.} $K$ is finite for the die (24 elements) and for
$S_5$ (120), and a continuous rotation group if the symbols are generic
rotations. Because $K$ is compact, a continuous injective $\iota$ is a
homeomorphism onto its image, so the encoding loses nothing. ``Every
length'' separates learning the rule from fitting the training lengths.
Continuity of the readout separates learning from memorization: two generic
unit quaternions already generate infinitely many distinct states, and a
readout allowed to jump could label each state with its answer, for any
group. A continuous readout cannot, since it would send arbitrarily close
states to different answers. Continuity is the minimal assumption that rules
out memorization, and every neural readout satisfies it.

\emph{Proof of Theorem~\ref{thm:main}.} (i) Let $R=\{Q(w)\}$ be the
reachable states, including $Q(\varnothing)=1$. Since
$Q(ww')=Q(w')Q(w)$, $R$ is a monoid, and its closure $C$ is a closed
submonoid of $A^\times$, compact by assumption, hence a compact group by
Lemma~\ref{lem:monoid}. If $Q(w)=Q(w')$ then
$\iota(\pi(w))=\rho(Q(w))=\rho(Q(w'))=\iota(\pi(w'))$, so $\pi(w)=\pi(w')$:
the state determines the answer. Because $K$ is compact, $\iota$ is a
homeomorphism onto the closed set $\iota(K)$, and $\rho(C)\subset
\overline{\rho(R)}\subset\iota(K)$. Hence $\varphi=\iota^{-1}\circ\rho$ is a
continuous map $C\to K$ with $\varphi(Q(w))=\pi(w)$. On $R$,
$\varphi(Q(w')Q(w))=\pi(ww')=\pi(w')\pi(w)$; both sides of
$\varphi(yx)=\varphi(y)\varphi(x)$ are continuous on $C\times C$ and agree
on the dense set $R\times R$, so $\varphi$ is a homomorphism. Its image is
compact and contains every answer, so it is $K$. With $N=\ker\varphi$, a
closed normal subgroup, the induced map $C/N\to K$ is a continuous
bijection from a compact space to a Hausdorff one, hence an isomorphism of
topological groups.

(ii) Let $p\colon C\to K$ be the quotient map followed by the isomorphism
$C/N\cong K$, and choose transitions $a(s)\in p^{-1}(g(s))$. Since $p$ is a
homomorphism, $p(Q(w))=\pi(w)$ for every word. As a closed subgroup of the
Lie group $A^\times$, $C$ is a compact Lie group, so $K\cong C/N$ is one
too and has a faithful finite-dimensional representation, which gives a
continuous injective $\iota\colon K\to\RR^d$. The map $\iota\circ p$ is
continuous on the compact, hence closed, set $C\subset A$; by the Tietze
extension theorem, applied to each coordinate, it extends to a continuous
$\rho\colon A\to\RR^d$, and $\rho(Q(w))=\iota(\pi(w))$.

(iii) If every element of $N$ is a scalar multiple of $1$ (real, or
complex when $A$ is a complex algebra), its elements commute with $C$, so
$C$ is a central extension of $K$; a finite central kernel makes it a cover.
Left multiplication by $C$ is a linear representation on $A$ in which $N$
acts by scalars, so it induces a homomorphism from $K\cong C/N$ to the
projective group of $A$, a projective representation of $K$. $\square$

\emph{Margin version.} Exact equality is a convenience. For finite $K$, say
a channel tracks $K$ \emph{with margin} if there are pairwise disjoint
closed sets $D_k\subset\RR^d$, one per $k\in K$, and a continuous readout
with $\rho(Q(w))\in D_{\pi(w)}$, as an argmax classifier with a positive
margin provides. Theorem~\ref{thm:main} holds unchanged. Indeed,
$\rho(C)\subset\bigcup_kD_k$ since the union is closed; the sets
$\rho^{-1}(D_k)\cap C$ are finitely many, closed and disjoint, so each is
also open in $C$, and $\varphi(x)=k$ for $\rho(x)\in D_k$ is locally
constant, hence continuous; it agrees with $\pi$ on $R$ because the $D_k$
are disjoint; and the homomorphism property follows by density as before.
The converse is immediate from (ii).

\emph{The readout chooses what is meaningful.} $N$ is the part of the state
the readout ignores. The 24-hour clock read as a 12-hour clock is the model
case. For quaternions, $q$ and $-q$ are the same rotation, so a quaternion
memory of rotations carries a sign a good readout ignores. The simplest such
readout is quadratic, $\rho(q)=L(qq^{\mathsf T})$, since
$(-q)(-q)^{\mathsf T}=qq^{\mathsf T}$; for matrix states, products
$Q_{ij}\overline{Q_{kl}}$ play the same role. Covers make small memories
possible: a group whose faithful representations need more dimensions may
still fit in fewer through a cover, provided the readout is invariant to
the extra bit, as features in an equivariant network are invariant to a
nuisance transformation.

\section{Proofs of Proposition~\ref{prop:reduction} and Corollary~\ref{cor:four}}\label{sm:reduction}

\emph{Intuition.} A symmetric matrix can be rotated into a diagonal one,
splitting into one-dimensional pieces. A memory can likewise be rotated into
independent blocks, but because its transitions need not commute, the
blocks cannot all be one-dimensional.

\emph{Proof of Proposition~\ref{prop:reduction}.} Since $C$ is closed under
products and contains $1$, its span $B$ is a subalgebra, and
$C\subset B^\times$ because $C$ contains the inverses of its elements. The
states lie in $C\subset B$, and the readout restricts to $B$, so the
channel regarded as a channel over $B$ still tracks $K$. Averaging an inner
product on $A$ over the compact group $C$ makes left multiplication by
elements of $C$ orthogonal, so every subspace of $A$ preserved by $C$, and
hence by $B$, has a preserved orthogonal complement: $A$ is a completely
reducible $B$-module. It is faithful, since $b\cdot1=b$. The Jacobson
radical of $B$ annihilates every simple module, hence every completely
reducible one, hence $A$, so it is zero and $B$ is semisimple.
Equation~\eqref{eq:wedderburn} is the Wedderburn--Artin theorem combined
with Frobenius's theorem that $\RR$, $\CC$ and $\HH$ are the only real
division algebras~\cite{lam}. Multiplication in a product of algebras is
componentwise, so the factors evolve independently. The projection of $C$
to a factor $M_{n_i}(D_i)$ is a compact subgroup of $GL_{n_i}(D_i)$;
averaging a $D_i$-Hermitian form on $D_i^{n_i}$ over it gives a basis in
which it lies in $O(n_i)$, $U(n_i)$ or $Sp(n_i)$. This parallels the
splitting of a representation of a compact group into irreducible
blocks~\cite{broecker}. $\square$

\emph{Proof of Corollary~\ref{cor:four}.} By
Proposition~\ref{prop:reduction}, $\dim B\le\dim A$. Since $\dim
M_n(D)=n^2\dim D$, the semisimple algebras of dimension at most three are
products of copies of $\RR$ and $\CC$, which are commutative; then $C$ is
abelian and so is its quotient $K$. In dimension four the semisimple
algebras are $\RR^4$, $\RR^2\times\CC$, $\CC^2$, $M_2(\RR)$ and $\HH$, of
which only the last two are non-commutative, and both track non-abelian
groups (SM~\ref{sm:ladder}). $\square$

\emph{What ``numbers'' counts.} Corollary~\ref{cor:four} counts the
dimension of the algebra, which is what the parallel scan multiplies. A
matrix channel can store less: left multiplication acts on each column
separately, so it may keep a single column $v_k=Q(w)v_0$.

\begin{lemma}[Column states]\label{lem:column}
Let $C\subset GL_n(D)$ be compact and $v_0\in D^n$ have trivial stabilizer in
$C$. Then the channel storing $v=Qv_0$ tracks exactly the groups that the
full channel storing $Q$ tracks. For finite $C$, a generic $v_0$ has trivial
stabilizer.
\end{lemma}
\emph{Proof.} The orbit map $C\to Cv_0$, $Q\mapsto Qv_0$, is a continuous
bijection from a compact space to a Hausdorff one, hence a homeomorphism,
and it maps the reachable states $Q(w)$ to the stored columns $Q(w)v_0$. A continuous readout of $Q$
on $C$ is thus a continuous readout of $Qv_0$ on the compact orbit, and
conversely; both extend to the whole space by Tietze. For finite $C$, each
$c\ne1$ fixes only the proper subspace $\ker(c-1)$, and finitely many proper
subspaces do not cover $D^n$. $\square$

So a two-number state with $2\times2$ real transitions tracks the dihedral
groups; what it cannot avoid is composing $2\times2$ transitions in the
scan. A quaternion channel cannot be shrunk this way, since $\HH$ has no
smaller faithful module, so for $\HH$ state and transitions both take four
numbers.

\section{The algebra ladder in detail}\label{sm:ladder}

By Theorem~\ref{thm:main}, a channel over $A$ tracks $K$ exactly when $K$
is a quotient $C/N$ of a compact subgroup $C$ of $A^\times$ by a closed
normal subgroup. The ladder lists these quotients for the small algebras.

\emph{One real number.} The compact subgroups of $\RR^\times$ are $\{1\}$
and $\{\pm1\}$, so a real channel tracks the trivial group or $C_2$, parity.
Positive real channels (Mamba-1, Mamba-2) have only $\{1\}$ and track
nothing; allowing $-1$ adds one bit~\cite{grazzi}.

\emph{Complex numbers.} The compact subgroups of $\CC^\times$ are the groups
$\mu_n$ of $n$-th roots of unity and the circle $U(1)$; their quotients are
cyclic groups and the circle. A complex channel therefore tracks exactly
counting modulo $n$ and continuous phases. A layer of $N$ complex channels
is one channel over $\CC^N$, still commutative: it tracks every finite
abelian group with enough channels, as a product of cyclic groups, and no
other finite group, at any width.

\emph{$2\times2$ real matrices.} Averaging an inner product conjugates every
compact subgroup of $GL_2(\RR)$ into $O(2)$, whose closed subgroups are
the cyclic groups $C_n$, the dihedral groups $D_n$ (the $2n$ symmetries of
a regular $n$-gon), $SO(2)$ and $O(2)$. A channel over $M_2(\RR)$ therefore
tracks the dihedral groups and their quotients, the first non-abelian
groups on the ladder.

\emph{Quaternions.} The norm is multiplicative and the only compact subgroup
of $\RR_{>0}$ is $\{1\}$, so every compact subgroup of $\HH^\times$ lies in
the unit quaternions $SU(2)$. The closed subgroups of $SU(2)$ are, up to
conjugacy, the cyclic groups, the binary dihedral groups $2D_n$ (with
$Q_8=2D_2$), the binary tetrahedral, octahedral and icosahedral groups
$2T$, $2O$, $2I$ of orders 24, 48 and 120, the circle $U(1)$, its normalizer
$U(1)\cup\mathbf jU(1)$, and $SU(2)$ itself~\cite{conway_smith,broecker}.
Each finite one contains $-1$ except the cyclic groups of odd order, and
modulo $\pm1$ they give the rotation groups of 3D space: $C_n$, $D_n$, and
the rotations of the Platonic solids, $A_4$ (tetrahedron), $S_4$ (cube,
die) and $A_5$ (icosahedron). A quaternion channel therefore tracks $Q_8$
and the binary groups with a readout that reads the sign, and $D_n$,
$A_4$, $S_4$, $A_5$ and $SO(3)$ with one that ignores it. $A_5$ is not
itself a subgroup of $SU(2)$, which contains a single element of order two,
so it can only be tracked through its cover $2I$.

\emph{$S_5$ needs more.} We show that no layer of complex and quaternion
channels side by side tracks $S_5$, however wide. Say that a finite group
$F$ is a \emph{section} of a compact group $G$ if $F\cong H/M$ for a closed
subgroup $H\le G$ and a closed normal subgroup $M\trianglelefteq H$. By
Theorem~\ref{thm:main}, a channel over $A$ tracks $F$ only if $F$ is a
section of the compact group $A^\times$; for $A=\CC^m\times\HH^n$ the
compact subgroups lie in $U(1)^m\times SU(2)^n$.

\begin{lemma}\label{lem:s5section}
If $S_5$ is a section of a product $G_1\times G_2$ of compact groups, it is
a section of $G_1$ or of $G_2$.
\end{lemma}
\emph{Proof.} Let $S_5\cong H/M$, and let $H_1=H\cap(G_1\times1)$, a closed
normal subgroup of $H$, which we identify with a subgroup of $G_1$. Its
image in $H/M$ is a normal subgroup of $S_5$, hence $1$, $A_5$ or $S_5$.
If it is $S_5$, then $S_5\cong H_1/(H_1\cap M)$ is a section of $G_1$. If it
is $1$, then $H_1\subset M$ and $S_5$ is a quotient of $H/H_1$, which is
isomorphic to the projection $p_2(H)$, a closed subgroup of $G_2$. If it is
$A_5$, let $L=H_1\cap M$, so that $H_1/L\cong A_5$. Conjugation by
$h=(h_1,h_2)\in H$ acts on $H_1\subset G_1\times1$ through $h_1$ alone, and
it preserves $H_1$ and $L$; so the compact group $P=p_1(H)\le G_1$
normalizes $H_1$ and $L$ and acts on $H_1/L\cong A_5$ by automorphisms. This
action agrees with the conjugation action of $H/M\cong S_5$ on its normal
subgroup $A_5$, which realizes every automorphism of $A_5$, since
$\operatorname{Aut}(A_5)\cong S_5$. The kernel $Z$ of $P\to\operatorname{Aut}(A_5)$
is closed (it is cut out by finitely many conditions, one per element of
$H_1/L$), so $S_5\cong P/Z$ is a section of $G_1$. $\square$

\begin{lemma}\label{lem:su2}
$S_5$ is not a section of $U(1)$ or of $SU(2)$.
\end{lemma}
\emph{Proof.} Sections of $U(1)$ are abelian. A finite quotient $H/M$ of a
closed subgroup $H\le SU(2)$ has $M$ open in $H$, so $M$ contains the
identity component of $H$. For $H=SU(2)$ or $U(1)$ the quotient is then
trivial, and for the normalizer of $U(1)$ it has order at most two. For
finite $H$, $|H/M|=120$ requires $|H|\ge120$; the cyclic and binary dihedral
groups are solvable, and the only other candidate is $2I$, of order 120,
whose quotients are $2I$, $A_5$ and $1$, none of them $S_5$ ($2I$ has a
single element of order two, $S_5$ has many). $\square$

By induction on the number of factors, Lemmas~\ref{lem:s5section}
and~\ref{lem:su2} show that $S_5$ is not a section of
$U(1)^m\times SU(2)^n$: no width of complex and quaternion channels
tracks it [Fig.~\ref{fig:intuition}(f)]. Concurrent work~\cite{jia_ncssm} reaches a
related exclusion through group laws satisfied by products of $SO(3)$ and
$SU(2)$ heads; the argument here uses only the outer automorphism of $A_5$. $S_5$ does fit in $M_2(\HH)$,
sixteen real numbers: it embeds in $SO(5)$ as the permutation action on the
four-dimensional space orthogonal to $(1,\dots,1)$ together with the sign
on a fifth axis, which has determinant one; its preimage in
$Spin(5)\cong Sp(2)$, the compact group of unitary $2\times2$ quaternion
matrices, is a double cover read
with a sign-invariant readout. It also fits, without a cover, in $4\times4$
real matrices, which permute the five corners of a regular 4-simplex. With
depth, five numbers suffice (Proposition~\ref{prop:depth}): a sign channel
beneath a quaternion channel.

Covers keep transitions small: through its cover $2I$, $A_5$ is tracked by
one quaternion, four numbers per transition and of state, while the smallest
faithful real representation of
$A_5$ needs $3\times3$ matrices (9 numbers per transition, 3 of state) and
the permutation representation $5\times5$ ones (25 per transition, 5 of
state)~\cite{pdssm}.

\section{Proof of Proposition~\ref{prop:depth}}\label{sm:depth}

\emph{Example.} For the six symmetries of a triangle, a first layer tracks
whether an odd number of reflections has occurred (a single sign). A second,
complex layer tracks the rotation, turning by $+120^\circ$ or $-120^\circ$
according to that sign, since after a reflection every later rotation is
seen in a mirror. Each layer commutes; the stack does not. The same works for
any group that can be taken apart one commutative piece at a time: the
solvable groups, whose derived length counts the pieces.

\emph{Setup.} Layer $\ell$ has state $x^\ell$ in a commutative algebra
$A_\ell$ and transitions in a compact abelian group $C_\ell\subset
A_\ell^\times$; reading $s_k$ multiplies $x^\ell$ by
$a_\ell(s_k,x^{<\ell}_{k-1})$, where $x^{<\ell}$ collects the states of the
layers below. Let $K=K^{(0)}\trianglerighteq K^{(1)}\trianglerighteq\cdots$
be the derived series, $K^{(i)}=[K^{(i-1)},K^{(i-1)}]$.

\emph{Proof of ``if''.} Let $K^{(L)}=1$. For $i=1,\dots,L$ fix a set $T_i$
of representatives of $K^{(i-1)}/K^{(i)}$ in $K^{(i-1)}$, containing $e$.
Every $g\in K$ is uniquely $g=t_1t_2\cdots t_L$ with $t_i\in T_i$. Reading
$s$ sends $g$ to $sg$: write $st_1=t_1'n_1$ with $t_1'\in T_1$,
$n_1\in K^{(1)}$, then $n_1t_2=t_2'n_2$, and so on, so that
$sg=t_1'\cdots t_L'$. In the abelian group $K^{(i-1)}/K^{(i)}$, the class of
$t_i'$ is the class of $t_i$ multiplied by the class of $n_{i-1}$ (with
$n_0=s$), and $n_{i-1}$ depends only on $s$ and $t_1,\dots,t_{i-1}$. Layer
$i$ stores the class of $t_i$ in the finite abelian group
$K^{(i-1)}/K^{(i)}$, represented faithfully by roots of unity in a few
complex channels, and multiplies it by the image of $n_{i-1}$, a function of
the symbol and of the previous states of the layers below. All layers have
finitely many reachable states, and the readout maps the joint state to
$\iota(t_1\cdots t_L)$, a function on a finite set, extended continuously.
Hence the stack tracks $K$.

\emph{Proof of ``only if''.} Let $Y_\ell$ be the finite set of reachable
joint states of layers $1,\dots,\ell$ for $\ell<L$. Each symbol acts on
$Y_{\ell}$ by an injective map, since every layer multiplies by an
invertible element given the layers below, so by a permutation of the
finite set $Y_\ell$. Let $G_\ell$ be the group of permutations of $Y_\ell$
generated by the symbols, and $G_L$ the closure of the monoid generated by
the symbols acting on $Y_{L-1}\times A_L$, inside the compact group of maps
$(y,z)\mapsto(\sigma y,f(y)z)$ with $\sigma\in G_{L-1}$ and $f\colon
Y_{L-1}\to C_L$. By Lemma~\ref{lem:monoid}, $G_L$ is a compact group.
Forgetting the top layer is a homomorphism $G_\ell\to G_{\ell-1}$ whose
kernel consists of maps $(y,z)\mapsto(y,f(y)z)$, which multiply pointwise
in the abelian group $C_\ell$ and so form an abelian group. By induction,
$G_\ell$ is solvable of derived length at most $\ell$; for $\ell=1$, $G_1$
is a quotient of a subgroup of $C_1$, abelian. The state after a word is the
image of the initial state under the corresponding element of $G_L$, a
continuous function of that element, so the argument of
Theorem~\ref{thm:main}(i) applies with $G_L$ in place of $C$: the answer
defines a continuous homomorphism from $G_L$ onto $K$. A quotient of a
group of derived length at most $L$ has derived length at most $L$.
$\square$

\emph{Why the finiteness assumption.} The argument needs a compact group of
joint transformations before taking the quotient. If the layers below take a
continuum of values, the maps $f$ of the upper layers range over functions
of a continuous state, and the closure of those generated need not be
compact; a group onto which a mere submonoid maps need not be solvable (free
monoids embed in metabelian groups). Whether the bound holds without the
assumption is open. It holds in every model in this paper, where the lower
layers are sign or snapped complex channels.

\emph{Consequences.} $Q_8$ and $A_4$ have derived length two and $S_4$
three. $A_5$ and $S_5$ are not solvable. The bound agrees with
circuit-complexity results~\cite{merrill_illusion,sarrof}: the word problems
of $A_5$ and $S_5$ are NC$^1$-complete~\cite{barrington}, while those of
solvable groups are not believed to be. The two-layer $S_5$ model is the
same construction with a non-commutative top layer: the sign channel tracks
$S_5/A_5$, and the quaternion layer multiplies by a lift of
$n_1=t_1'^{-1}st_1\in A_5$ into $2I$, read with a sign-invariant readout.
Two-layer semidirect cascades of this kind for $S_5$ also appear
in concurrent work~\cite{jia_ncssm}; ours uses one real and one quaternion channel, five
numbers in all.

\section{HyperSSM: parametrization, cost and stacking}\label{sm:hyperssm}

\emph{The exponential.} Writing $a=e^{\Delta(-\alpha+v)}$ splits a step into
fading ($-\Delta\alpha$) and turning ($\Delta v$), as SSMs write
$a=e^{\Delta\lambda}$. $\alpha=0$ gives pure turning, the tracking regime;
raising $\alpha$ lets a channel forget. For complex channels, $v=i\omega$
gives $a=e^{-\Delta\alpha}e^{i\Delta\omega}$, exactly a diagonal SSM channel.
For quaternion channels, $v=v_1\mathbf i+v_2\mathbf j+v_3\mathbf k$ squares
to $-|v|^2$, and
\begin{equation}
a = e^{-\Delta\alpha}\bigl[\cos(\Delta|v|) + \sin(\Delta|v|)\,v/|v|\bigr].
\label{eq:quat}
\end{equation}
Without decay this is a 3D rotation by angle $2\Delta|v|$ about $v/|v|$. If
$v$ is confined to one axis, all transitions commute and the channel is
complex again. With skew-Hermitian $v$ the same recipe reaches every bounded
transition of the complex and quaternion blocks of
Proposition~\ref{prop:reduction}, whose unitary groups are connected.

\emph{Real channels.} The exponential of a skew-symmetric matrix is a
rotation, never a reflection, so the transition $-1$ of a real channel, and
every odd permutation in $M_n(\RR)$, are out of its reach. HyperSSM gives
real channels a learned sign. A smooth relaxation such as
$a=\tanh(g)$ does not work: the bounded real transitions $\{\pm1\}$ are two
points, so changing a sign means passing through $|a|\approx0$, where the
state is erased, and any $|a|<1$ decays over long sequences. A real channel
is therefore exactly $a=\operatorname{sign}(g)$ in the forward pass, in
training and at test, with the gradient of $\tanh(g)$ in the backward pass
(a straight-through estimator). This uses nothing about the task, since
$\pm1$ are the only bounded real transitions; it plays the role of the
hardmax of PD-SSM~\cite{pdssm}. Complex and quaternion transitions need no
such step: their bounded transitions form connected groups, which a
continuous parametrization covers with no decay.

\emph{Cost and readout.} Steps combine by Eq.~\eqref{eq:combine}, so all
prefix states follow from the same scan; a quaternion product costs 16 real
multiplications against 4 for a complex one.
HyperSSM offers a \emph{raw} readout of $x$ and a \emph{sign-invariant}
readout of the ten entries of $xx^{\mathsf T}$ for a quaternion channel. Cheap complex channels serve
counting, position and periodicity; quaternion or matrix channels serve
non-commuting operations. A real sign channel beneath a quaternion layer
tracking $A_5$ tracks $S_5$ with 5 numbers of state rather than 16, by the
construction of Proposition~\ref{prop:depth}.

\emph{Stacking, inputs and dense blocks.} (i) In a stack, the generator of a
layer is bilinear in the input and in the \emph{characters} of the layer
below at the previous step,
$v_k=W\bigl(u_k\otimes[1,\chi(x^{\mathrm{below}}_{k-1})]\bigr)$, where
$\chi$ is the readout's feature map: $(\mathrm{Re}\,x^m,\mathrm{Im}\,x^m)$
for complex channels and $xx^{\mathsf T}$ for sign-invariant quaternion
ones. This is the transition of Proposition~\ref{prop:depth}; since
characters span the functions on a finite group, it is fully expressive while
staying linear. (ii) Transitions may depend on the previous token as well as
the current one, as through Mamba's short convolution. (iii) A channel over
$M_n(\CC)$ holds a vector of $\CC^n$ and multiplies it by a general complex
matrix, which may be singular; it is the costliest rung, used only where a
smaller one does not train. With channels over $\RR$, $\CC$ and $\HH$, a
HyperSSM layer keeps the $O(T)$ work, $O(\log T)$ depth and constant memory
per token of a diagonal SSM; a dense $M_n(\CC)$ block adds a factor $n^3$ to
the scan.

\section{Benchmark protocol, training and full results}\label{sm:bench}

\emph{Protocol.} Following~\cite{walker_slice,pdssm}: training sequences of
length 3 to 40, loss on the final token only, validation on lengths 40 to
256, best checkpoint. Tasks: the four regular languages of~\cite{deletang}
(parity; even pairs; navigation on a cycle of five positions; arithmetic
modulo 5 with multiplication before addition and subtraction) and the word
problems of $A_5$ and $S_5$ with $n$ generators: two standard ones and $n-2$
random group elements, which raise the connectivity of the
automaton~\cite{pdssm}. On the group word problems we also test
on lengths 256 to 1000, 25 times the training length, on fresh
sequences not used for checkpoint selection. Baseline numbers are those
reported in~\cite{pdssm}; its extra generators are random and unpublished,
so our $A_5$ and $S_5$ tasks match in size but not exactly.

\emph{Model choice.} One real channel with a sign for parity ($\ZZ_2$, whose
irreducible representation is real); one complex channel for cycle
navigation ($\ZZ_5$); one complex channel on pairs of consecutive tokens for
even pairs, since the first and last symbols agree exactly when the number
of changes between neighbours is even; one quaternion channel with the
sign-invariant readout for $A_5$; a real sign channel beneath a quaternion
channel for $S_5$. Readouts are linear in the characters of the top layer.

\emph{Nothing about the task's group is supplied.} Real channels
are $\pm1$ in the forward pass in training and at test. Complex channels are
trained continuously. Training on lengths up to 40 pins a phase only to
within about the gap between answers divided by 40, so a continuous phase
that is right on the training lengths drifts beyond them. By
Theorem~\ref{thm:main} a tracking channel's transitions form a finite group,
and the finite groups of unit complex numbers are the $n$-th roots of unity.
Each complex channel is therefore snapped to the \emph{smallest} $n$
(searched up to 64) whose roots of unity give the best accuracy on the
validation lengths, the set already used for checkpoint selection; $n$ is
read off the trained model, never given. The same snapped score selects
checkpoints during training, and after the snap the readout is fine-tuned on
the snapped states, so the model evaluated is the model last trained.
Quaternion channels and the dense $M_{16}(\CC)$ block are continuous
throughout, and no transition is rounded at evaluation.

\emph{What training needs.} First, \emph{a minimal channel with a linear
readout}. $K$ phase channels read linearly give $2K$ Fourier features of the
sequence, which can fit any labelling of the lengths up to $T$ once
$2K+1>T$: a wide or nonlinear readout can memorize the training lengths
instead of learning the group. With sixteen complex channels and a small
MLP, parity fit its training data and stayed at chance beyond it; one real
channel with a linear readout learned it exactly. This is
Definition~\ref{def:tracking} at work. Second, \emph{a short-length
curriculum} within the protocol: lengths 3 to 5 for the first 30\% of
training, then a ramp to 40. Third, \emph{restarts}: a run either finds the
group or stays near chance, so an attempt still far from exact halfway
through training is restarted, up to 8 times for $A_5$ and $S_5$ and 20
times for the one-channel tasks.
No layer is pretrained or given extra supervision: for $S_5$, the
sign channel finds the abelianization $S_5\to\{\pm1\}$ of
Proposition~\ref{prop:depth} from the final answers alone, trained together
with the quaternion layer above it. This works because the sign is exactly
$\pm1$ in the forward pass; with $a=\tanh(g)$, the signs stay where they
start.

\emph{Additional results.} The snap found $n=2$ for even pairs and
$n=5$ for cycle navigation in every run, without being told either. Before
the snap, the same trained models score near chance beyond the training
lengths, at 49\% and 17\% on lengths 40--256: they had learned the group,
and the continuous phases hid it. On the group tasks, the
quaternion channel is continuous and the sign and quaternion layers of $S_5$
are trained together. PD-SSM is exact up to 8 generators of $S_5$. The best
run of every group task stays exact on lengths 256--1000. Continuous
channels drift slowly in some other runs (mean over runs on 256--1000:
94.8--99.9\% for $A_5$, 92.3\% for $S_5$ with 32 generators); not every run
of $S_5$ with 8 generators finds the group.

\subsection{Modular arithmetic: representation versus optimization}\label{sm:modarith}

The automaton of modular arithmetic is not a group, but its pieces are
simple: the nonzero residues under multiplication form $\ZZ_4$, generated by
2; sums form $\ZZ_5$; and multiplying by zero or starting a new term is a
reset~\cite{krohn_rhodes}. Two commutative layers suffice: a complex channel
tracking the current product, reset at each new term and sent to zero by a
factor 0, beneath a complex channel tracking the sum and turning, at each
$+$, $-$ or $=$, by an angle read linearly from the characters of the first.
This is three real numbers of state, and, set by hand, it is exact on every
length up to 1000. To learn it, training must discover the discrete
logarithm mapping each residue to its rotation, the two layers must agree on
it, and the absorbing zero dominates the training signal. Here
earlier models need large dense blocks: in~\cite{walker_slice},
block-diagonal transitions improve as blocks grow, from 37.7\% with blocks
of two to 76.6\% with blocks of sixteen, and the best structured model adds
a dense $16\times16$ block. One dense $M_{16}(\CC)$ channel on pairs of
consecutive tokens reaches $99.3\pm0.4\%$ on lengths 40--256 over
5 runs, above PD-SSM ($96.2\pm3.4\%$) and second only to the LSTM
($99.9\pm0.1\%$). Unlike the smaller channels, its transitions are general
complex matrices, with no finite group to snap to. A small exact memory
exists; finding it by gradient descent is open.

\section{Gyroscope experiment: setup}\label{sm:imu}

A gyroscope reports the angular velocity $\omega_k$ of a body in its
own frame, and the rotation between two times is
$\Delta q=e^{\omega_1\Delta t/2}\cdots e^{\omega_T\Delta t/2}$ in
quaternions. Composing the same measured rotations in reverse order gives a
median error of $41^\circ$ after 2\,s, against $1^\circ$ in the right order.

\emph{Data.} BROAD~\cite{broad}: 39 recordings of one inertial
sensor at 286\,Hz with orientation from optical motion capture, averaged to
57\,Hz, split into training (28), validation (3) and test (8); the test
recordings are those that rotate. A model reads only the gyroscope readings
in a window and predicts the rotation over it, with loss
$1-\langle\hat q,q\rangle^2$ on the final step, which ignores the sign of a
quaternion. Training windows are 1--2\,s; test windows 1\,s to 2\,min.

\emph{Models.} HyperSSM is one quaternion channel with generator
$v_k=W\omega_k+b$, 12 parameters in total. Baselines see the same inputs and
are trained the same way: a selective complex diagonal SSM with 64 channels
and an MLP readout (the memory of Mamba-3), and an LSTM with 128 units. Two
untrained references: integrating the raw readings, and predicting no
rotation.

\emph{Detailed results.} Complex diagonal SSM vs no rotation:
$32.5^\circ$ vs $43.6^\circ$ at 1\,s, and no better at every longer length.
LSTM: $3.2^\circ$ at 1\,s, $32^\circ$ at 10\,s, worse than no rotation from
30\,s on. HyperSSM vs integration: $7.5^\circ$ vs $8.6^\circ$ at 30\,s,
$26.5^\circ$ vs $30.9^\circ$ at 2\,min, spread at most $2.5^\circ$ over
seeds. $W$ and $b$ are the sensor's scale, axis misalignment and bias, and
can be read off directly. Nothing in the architecture is specific to
inertial sensing: the general HyperSSM layer with the algebra the theory
prescribes is the physical integrator.

\section{Qubit experiment: setup}\label{sm:qubit}

\emph{Task.} A qubit starts in $|0\rangle$; a random sequence of
gates from $\{H,S\}$ (Clifford) or $\{H,S,T\}$ (Clifford+$T$) is applied,
and the label is the Bloch vector $r=(\langle X\rangle,\langle Y\rangle,
\langle Z\rangle)$ of the final state, from which the Born rule gives every
outcome probability, $P(\pm)=(1\pm r_i)/2$. As Bloch-sphere rotations, $H$
turns by $\pi$ about $(\hat x+\hat z)/\sqrt2$, $S$ by $\pi/2$ about $\hat z$
and $T$ by $\pi/4$ about $\hat z$; global phases are irrelevant. Labels were
checked against an explicit simulation with $2\times2$ unitaries (agreement
to $2\times10^{-8}$). Training circuits have 1--40 gates, with 1--5 gates
for the first 30\% of training; test circuits have 1 to 1000 gates.

\emph{Imperfect gates.} Each gate is replaced, once and for all, by
a rotation whose angle is $2\%$ short and whose axis is tilted by
$2^\circ$ in a random direction. Labels use these gates; the models see only
the gate names; the reference simulator uses the ideal gates.

\emph{Models.} HyperSSM is one quaternion channel in the tracking
regime, $x_k=e^{v_{g_k}}x_{k-1}$ with one learned $v_g\in\operatorname{Im}\HH$
per gate, and a readout linear in the ten entries of $xx^{\mathsf T}$, which
gives the Bloch vector exactly since its components are quadratic in $x$.
The baselines are a three-layer selective complex diagonal stack (32
channels per layer, MLP blocks between layers), the depth
Proposition~\ref{prop:depth} requires for the Clifford group, and an LSTM
with 128 units. All models use the loss $\|\hat r-r\|^2$ and the best of 8
short starts by training loss. The error is the angle between $\hat r$ and
$r$.

\emph{Decoding the gates.} The readout fixes the Bloch sphere only
up to a common rotation, so the learned gates are compared with the true
ones through frame-independent quantities: each gate's rotation angle and
the angles between gate axes. In all three settings these agree with the
true gates to three decimal places, including the imperfect gates ($H$:
$176.400^\circ$, $S$: $88.200^\circ$, $T$: $44.100^\circ$; axes $H$--$S$
$46.643^\circ$, $H$--$T$ $44.580^\circ$, $S$--$T$ $2.071^\circ$).

\begin{figure}[!htb]
\centering
\includegraphics[width=\textwidth]{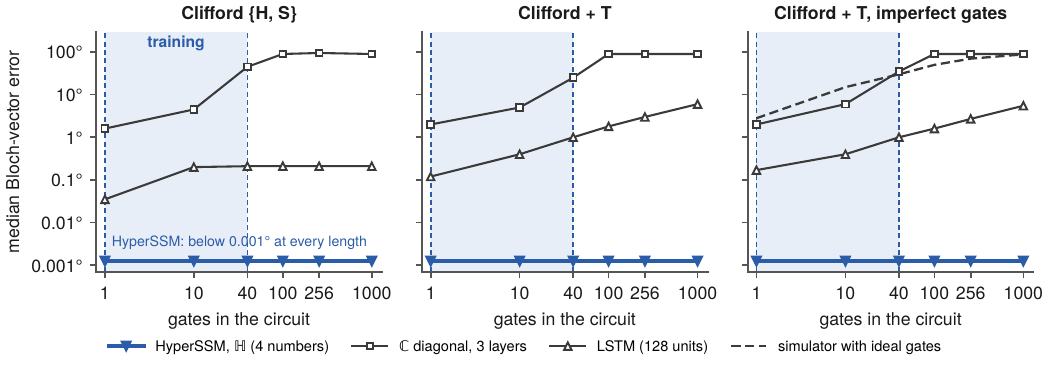}
\caption{\textbf{Qubit, all three settings.} Median angle between
predicted and true Bloch vectors against circuit length; training circuits
1--40 gates (shaded). HyperSSM is below the bottom of the axis ($10^{-3}$
degrees) at every length in every setting. With ideal gates the simulator is
exact and is not shown.}
\label{fig:qubit_sm}
\end{figure}

\section{Protein backbone experiment: setup}\label{sm:protein}

\emph{Data.} X-ray structures from the Protein Data Bank~\cite{pdb}
with resolution at most $1.8$\,\AA, one protein entity and 150--800
residues; 600 entries were sampled, of which 554 were parsed. Backbone atoms
N, C$^\alpha$, C are taken from the first model (alternate location blank or
A), and chains are cut into segments wherever the peptide bond C--N is
missing or longer than $1.7$\,\AA. Entries are split into training (442
entries, 147{,}661 residues), validation (55; 18{,}168) and test (56;
18{,}071).

\emph{Frames and target.} Residue $i$ has frame
$R_i=[e_1\,e_2\,e_3]$ with $e_1$ along C$^\alpha\!\to$C, $e_2$ in the
N--C$^\alpha$--C plane, and $e_3=e_1\times e_2$. The target is
$R_{i\to j}=R_i^{\mathsf T}R_j=F_iF_{i+1}\cdots F_{j-1}$, with
$F_k=R_k^{\mathsf T}R_{k+1}$. For rigid bond geometry,
\begin{equation}
F_k=K_0\,e^{\psi_k u_\psi/2}\,K_1\,e^{\omega_k u_\omega/2}\,K_2\,e^{\phi_{k+1}u_\phi/2}\,K_3,
\label{eq:backbone}
\end{equation}
with constant rotations $K_m$ and fixed unit axes $u$, and torsions
$\psi_k=\angle(\mathrm N_k,\mathrm C^\alpha_k,\mathrm C_k,\mathrm N_{k+1})$,
$\omega_k=\angle(\mathrm C^\alpha_k,\mathrm C_k,\mathrm N_{k+1},\mathrm C^\alpha_{k+1})$,
$\phi_{k+1}=\angle(\mathrm C_k,\mathrm N_{k+1},\mathrm C^\alpha_{k+1},\mathrm C_{k+1})$.

\emph{Tokens and model.} A stretch of $n$ steps becomes $6n+1$
tokens $(\tau_t,\alpha_t)$: for each step, a bond token, $\psi$, a bond token,
$\omega$, a bond token, $\phi$; then a final bond token; eight types in all,
with $\alpha_t=0$ for bond tokens. HyperSSM is the tracking-regime channel
\begin{equation}
x_t=e^{v_t}x_{t-1},\quad x_0=1,\qquad v_t=W\bigl(e_{\tau_t}\otimes[1,\alpha_t]\bigr)=W_{\tau_t,0}+\alpha_tW_{\tau_t,1},\qquad \hat q=\bar x_T,
\end{equation}
where the conjugate readout turns the left products of the recurrence into
the chain order of Eq.~\eqref{eq:backbone} (the sign of $W$ is learned).
Setting $W_{\text{torsion},0}=0$, $W_{\text{torsion},1}=u/2$ and
$e^{W_{\text{bond},0}}=K$ reproduces ideal geometry exactly, so the model is
exact for rigid geometry. The input term is zero: each token acts only
through its transition. The baselines see the same tokens with features
(one-hot type, $\alpha$, $\cos\alpha$, $\sin\alpha$): a selective complex
diagonal SSM with 64 channels and an MLP readout, and an LSTM with 128
units.

\emph{Training.} Loss $1-\langle\hat q,q_{i\to j}\rangle^2$, error
$2\arccos|\langle\hat q,q_{i\to j}\rangle|$, both blind to the sign of $q$.
Stretches of 1--16 residues, with 1--2 for the first 30\% of training; 3000
steps; best checkpoint by the median error on validation stretches of 64
residues. Products of rotations have local minima: fitted to exact
ideal-geometry targets, HyperSSM reaches $0.0^\circ$ to $0.3^\circ$ from 3 of 8
random starts and about $32^\circ$ from the others. Every model therefore
tries 8 initializations for 300 steps on 1--2 residues and keeps the one
with the lowest training loss.

\emph{References.} Ideal geometry rebuilds the backbone from the
torsions with the bond lengths and angles of Engh and Huber~\cite{engh_huber}
(N--C$^\alpha$ $1.458$\,\AA, C$^\alpha$--C $1.525$\,\AA, C--N $1.329$\,\AA;
N--C$^\alpha$--C $111.2^\circ$, C$^\alpha$--C--N $116.2^\circ$, C--N--C$^\alpha$
$121.7^\circ$) and reads off the relative frame; the wrong-order reference
multiplies its per-residue rotations in reverse.

\emph{Results} are in Table~\ref{tab:protein}; the s.d. over
seeds is at most $0.3^\circ$ for HyperSSM up to 64 residues.

\begin{table}[!htb]
\caption{\label{tab:protein}\textbf{Protein backbone.} Median
orientation error (degrees) of residue $j$ in the frame of residue $i$ against
the number of residues between them, on held-out proteins; mean over 3 seeds.}
\centering
\small
\begin{tabular}{lccccccccc}
residues & 1 & 2 & 4 & 8 & 16 & 32 & 64 & 128 & 256\\
\hline

no rotation        & 106.2 & 133.5 & 107.5 & 128.4 & 140.0 & 132.0 & 129.6 & 136.1 & 129.5 \\
ideal geometry     & 2.7 & 4.0 & 6.3 & 9.2 & 14.4 & 20.4 & 26.5 & 46.0 & 63.7 \\
ideal, wrong order & 2.7 & 38.3 & 58.2 & 86.7 & 95.6 & 99.7 & 108.5 & 119.6 & 131.5 \\
HyperSSM   & 2.6$\pm$0.0 & 3.7$\pm$0.0 & 5.9$\pm$0.0 & 8.3$\pm$0.1 & 12.1$\pm$0.2 & 17.8$\pm$0.2 & 23.7$\pm$0.3 & 36.2$\pm$1.0 & 49.3$\pm$0.3 \\
C diagonal & 8.5$\pm$2.2 & 25.9$\pm$8.2 & 104.6$\pm$27.9 & 122.0$\pm$12.6 & 126.5$\pm$5.1 & 135.9$\pm$0.5 & 135.0$\pm$1.6 & 131.0$\pm$2.9 & 130.4$\pm$3.0 \\
LSTM       & 7.7$\pm$0.8 & 14.8$\pm$1.5 & 106.2$\pm$28.3 & 126.5$\pm$8.7 & 132.0$\pm$5.3 & 136.9$\pm$1.1 & 132.8$\pm$2.7 & 131.3$\pm$2.2 & 132.9$\pm$1.9 \\
\end{tabular}
\end{table}

\section{Reading trained models: proof of Proposition~\ref{prop:read} and procedures}\label{sm:interp}

\emph{Why the weights are legible.} A probe trained on hidden states can
find information the model does not use, and can miss information stored
nonlinearly~\cite{li_othello}. In a tracking channel neither problem
arises. Theorem~\ref{thm:main} fixes what the state is (an element of $C$),
Proposition~\ref{prop:reduction} fixes its coordinates up to a change of
basis within each block, and the readout's invariance fixes what is ignored
($N$). Interpreting the model reduces to identifying a homomorphism
$C\to K$, a finite problem when $K$ is finite. This is the recurrent
counterpart of the universality observed in networks trained on group
operations, which were found to compose elements through the group's
representations~\cite{chughtai_universality}, and in networks trained on
modular addition, which were found to rotate a clock~\cite{nanda_grokking}.
There it was an empirical finding about what training happened to produce;
for a linear recurrent memory that tracks, Theorem~\ref{thm:main} says it
cannot be otherwise, and a complex channel is that clock by construction.

\emph{Proof of Proposition~\ref{prop:read}.} (i) The pair
$(Q(w),\pi(w))$ evolves deterministically: reading $s$ sends $(Q,g)$ to
$(a(s,Q)\,Q,\;g(s)\,g)$, where $a(s,Q)$ is the transition chosen by $s$ and,
in a stack, by the states of the lower layers held in $Q$. If states take
values in a finite set $S$, this is a finite automaton on $S\times K$, and
the pairs reached by nonempty words are found by breadth-first search from
$(1,e)$ in at most $|S|\,|K|$ steps. The model tracks $K$ at every length if
and only if the readout gives the correct answer on each of these pairs,
since every word of every length ends in one of them. For a single channel
whose transitions generate a finite group $C$, the reachable states lie in
$C$, because a submonoid of a finite group generated by its elements is the
subgroup they generate. (ii) Let $a_k$ be the trained transitions along a
word and $a^*_k$ those of the certified model, with
$\|a_k-a^*_k\|\le\varepsilon$ and all norms at most one, so that decay
($\alpha>0$) is allowed. Telescoping,
\begin{equation}
Q(w)-Q^*(w)=\sum_{k=1}^{T} a_T\cdots a_{k+1}\,(a_k-a^*_k)\,a^*_{k-1}\cdots a^*_1,
\label{eq:telescope}
\end{equation}
so $\|Q(w)-Q^*(w)\|\le T\varepsilon$. For $T\varepsilon<r$ the state lies in
the ball of radius $r$ around $Q^*(w)$, inside the decision region of the
correct answer by (i). For a stack, the same bound applies to the top
channel when the layers below are exact, as the sign channel is. $\square$

\emph{Guarantee versus observation.} The horizon $r/\varepsilon$ is a
guarantee, not a prediction of failure. Errors that are systematic, such as
one generator turning slightly too far, add linearly; errors that differ
across symbols partly cancel, so a run may stay exact well beyond
$r/\varepsilon$. Comparing the two over runs tests whether the slow drift
of the continuous runs (SM~\ref{sm:bench}) is explained by the distance of
their transitions from an exact group.

\emph{Decoding a quaternion transition.} A unit quaternion
$a=\cos\tfrac\theta2+\sin\tfrac\theta2\,\hat n$ is a rotation by
$\theta=2\arccos|a_0|$ about $\hat n$; the absolute value makes the reading
independent of the sign the readout ignores. In $A_5$, elements of order
$2$, $3$ and $5$ act on an icosahedron as rotations by $180^\circ$,
$120^\circ$, and $72^\circ$ or $144^\circ$, and the axes of any two elements
meet at an angle between symmetry axes of one icosahedron.

\emph{Distance to an exact model.} For analysis only, the permutation
$g(s)$ of each symbol is used; it is never given in training. Fix an exact
lift $a^*(s)$ of $g(s)$ into the 120 unit quaternions of the binary
icosahedral group, for each of the two inequivalent embeddings of $A_5$ in
the rotation group (they are exchanged by $\sqrt5\mapsto-\sqrt5$, which
swaps $72^\circ$ and $144^\circ$). Then
\begin{equation}
\varepsilon=\min_{U,\ \text{embedding},\ \text{signs}}\ \max_s\,
\bigl\|a(s)\mp U a^*(s)U^{-1}\bigr\|,
\label{eq:eps}
\end{equation}
minimized over unit quaternions $U$ (three parameters, from several random
starts). For $S_5$, the sign channel is compared with the parity of each
generator; the top channel has one transition per symbol and sign below,
and the exact model also depends on the coset representative $\tau$ of
Proposition~\ref{prop:depth}, one of 60 odd permutations, searched
exhaustively. The radius $r$ follows from the readout: for an argmax over
logits linear in $qq^{\mathsf T}$, with margin $\mu$ on the exact states,
$r\ge\mu/\Lambda$, where $\Lambda$ bounds the Lipschitz constant of the
logit differences on the unit ball.

\emph{Gyroscope.} The exact integrator is $v=\omega\,\Delta t/(2\Delta)$,
that is $W=\tfrac{\Delta t}{2\Delta}I$ and $b=0$ in $v_k=W\omega_k+b$ (Eq.~\eqref{eq:quat}).
The learned $W$ departs from it by a symmetric part (axis scale factors and
cross-axis coupling) and an antisymmetric part (a small rotation between
sensor and body frames), and $b$ is the gyroscope bias up to the same
scale. These can be compared with a calibration obtained independently of
the model, for example the mean reading during rest segments of the
recordings.

\section{Scope and extensions}\label{sm:outlook}

The theory covers linear recurrences, bounded transitions and exact
readouts. Most sequences also overwrite memory, and by the Krohn--Rhodes
theorem~\cite{krohn_rhodes} every finite automaton is built from groups and
resets, suggesting that stacks of such channels with resets can recognize
all regular languages; modular arithmetic (SM~\ref{sm:modarith}) is an
example. Many physical processes, such as rigid-body rotations, as
with the gyroscope, qubit gates, holonomies and braids of anyons, are
ordered products in the same groups. So are the frames of a
protein backbone, and with translations added, rigid motions in general,
which dual quaternions compose in a single associative algebra. The certificate of
Proposition~\ref{prop:read} extends to any model whose states, after
projection, take finitely many values; whether channels inside large hybrid
models stay close enough to an exact group to be read this way is open.

\end{document}